\documentclass[conference]{IEEEtran}
\IEEEoverridecommandlockouts
\usepackage{cite}
\usepackage{amsmath,amssymb,amsfonts}
\usepackage{algorithmic}
\usepackage{graphicx}
\usepackage{textcomp}
\usepackage{xcolor}
\usepackage{multirow}
\usepackage{multicol}
\usepackage{booktabs}
\usepackage{enumitem}
\usepackage{amsfonts}
\usepackage{symbols}
\usepackage{hyperref}
\usepackage{geometry}

\def\BibTeX{{\rm B\kern-.05em{\sc i\kern-.025em b}\kern-.08em
    T\kern-.1667em\lower.7ex\hbox{E}\kern-.125emX}}

\begin{document}

\newgeometry{top=2.54cm, left=1.91cm, right=1.91cm, bottom=1.91cm}

\title{Visual-Language-Guided Task Planning for Horticultural Robots\\
}

\author{Jose Cuaran$^{1*}$, Kendall Koe$^{1*}$, Aditya Potnis$^{1}$, Naveen Kumar Uppalapati$^{3}$, and Girish Chowdhary$^{1,2}$

\thanks{*Equal Contribution, The authors are with (1) the Siebel School of Computing and Data Science, (2) the Department of Agricultural and Biological Engineering  and (3) National Center for Supercomputing Applications at University of Illinois, Urbana-Champaign.
        }%
\thanks{{Correspondence to \tt\small \{jrc9,girishc\}@illinois.edu}}
}

\maketitle


\begin{abstract}
Crop monitoring is essential for precision agriculture, but current systems lack high-level reasoning. We introduce a novel, modular framework that uses a Vision Language Model (VLM) to guide robotic task planning by actively querying heterogeneous data sources—including enriched RGB camera feeds and 2D semantic occupancy maps—interleaved with robotic action primitives. We contribute a comprehensive benchmark for short- and long-horizon crop monitoring tasks in monoculture and polyculture environments. Our results show that while zero-shot VLMs perform robustly for short-horizon tasks (achieving 87\% success, comparable to human experts), success drops significantly to under 10\% for complex long-horizon, multi-target tasks. Despite this decline, task completion rates remain above 76 \% under noiseless conditions. Critically, the system degrades when relying on noisy semantic maps, demonstrating a key limitation in current VLM context grounding for sustained robotic operations. This work offers a deployable framework and critical insights into VLM capabilities and shortcomings for complex agricultural robotics.\\
\href{https://kendallkoe.com/Visual-Language-Guided-Task-Planning-for-Horticultural-Robots/}{https://kendallkoe.com/Visual-Language-Guided-Task-Planning-for-Horticultural-Robots/}.
\end{abstract}


\begin{figure*}[!t]  
    \centering
    \includegraphics[width=\textwidth]{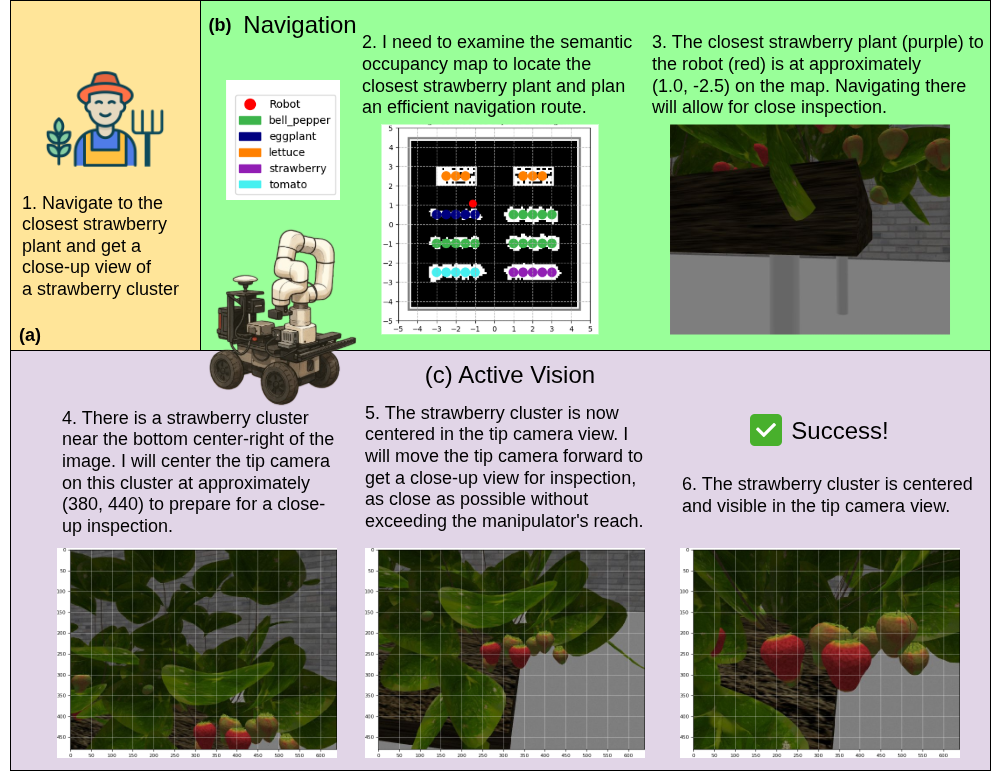}
    \caption{Sample episode of a single-plant, single-target monitoring task executed by the VLM agent. (a) The VLM takes as input the task prompt, the context and tools description. (b) The navigation tool call is used to navigate to the goal object and (c) The  manipulation tools are invoked at different time steps until completing the inspection task.}
    \label{fig:sample_episode}
\end{figure*}

\vspace{-0.1cm}
\section{Introduction}

The agricultural sector is under increasing pressure to meet the global demand for food while facing persistent labor shortages, climate variability, and sustainability challenges. Precision agriculture has emerged as a response, leveraging data-driven technologies to optimize resource use, improve productivity, and minimize environmental impact. Central to this approach is crop monitoring, which enables continuous assessment of plant health, early detection of pests and nutrient deficiencies, and precise management of irrigation and fertilization \cite{getahun2024application}. In recent years, robotics has become an essential component of precision agriculture, automating key tasks such as soil preparation, planting, weeding, harvesting, and phenotyping across diverse environments including fields, orchards, and greenhouses \cite{spagnuolo2025agricultural}. These advances pave the way for intelligent horticultural robots capable of not only perceiving and acting within complex crop environments but also reasoning about high-level monitoring tasks.

In horticultural environments, mobile manipulator robots show strong potential for active data collection and crop monitoring. Prior work has demonstrated their effectiveness for target-aware mapping and viewpoint planning \cite{cuaran2025active, pan2023panoptic, freeman2024autonomous, lehnert20193d, zaenker2021viewpoint}. However, most existing approaches focus on a single target type (e.g., leaves, fruits, or stems) within a single crop, and rely on a fixed mapping or sensing strategy. In real farming scenarios, monitoring needs are substantially more diverse. Growers may wish to perform different types of scans—such as dense 3D reconstruction of fruit clusters, sparse sampling for disease inspection, or targeted imaging of stems—across multiple plant parts and crop types within the same environment. Supporting this flexibility is critical, as such data underpins a wide range of downstream agricultural tasks, including precision pesticide application, harvest readiness assessment, and pruning or canopy management. Ideally, these capabilities should be accessible through intuitive human–robot interaction. Rather than configuring low-level sensing or planning objectives, a farmer should be able to specify high-level goals in natural language (e.g., “go to the tomato plants and collect 50 images of ripe fruits”), which the robot can interpret, decompose into subtasks, and execute autonomously. This paradigm shifts mobile manipulation from narrowly specialized sensing systems toward general-purpose agricultural assistants capable of adapting to evolving monitoring needs.

The rise of Large Language Models (LLMs) and Vision-Language Models (VLMs) has marked a paradigm shift for task planning in robotics, enabling zero-shot task decomposition from natural language instructions into low-level plans for navigation and active vision in different domains \cite{huang2022language, ahn2022can, huang2022inner, liang2022code}. VLMs extend this capability by directly interpreting visual inputs, supporting applications such as action scoring, object relationship extraction, success detection, and traversability assessment \cite{goetting2024end, nasiriany2024pivot, wu2024voronav, zhi2024closed, jiang2024roboexp, elnoor2024robot}, often augmented by maps for spatial context \cite{huang2022visual, shah2023lm, werby2024hierarchical, chen2024mapgpt, zhong2024topv}. Object-goal navigation and instruction following have been the main focus of these works.

Despite these advancements, the integration of LLMs and VLMs into agricultural robotics lags behind other domains. Applications in agriculture have primarily focused on disease and pest detection \cite{haghighat2025multimodal}, with limited exploration of task planning. Notable exceptions include AgriVLN \cite{zhao2025agrivln}, which employs VLMs for long-horizon navigation tasks in agricultural scenarios leveraging camera observations only. A concurrent work uses an LLM to generate behavior trees for mission planning using textual farm layout descriptions \cite{zuzuarregui2025leveraging}. However, these methods suffer from poor spatial reasoning as they rely on camera images or text descriptions only.

To address these gaps, we propose a VLM-driven framework for user-friendly crop monitoring with agricultural mobile manipulators. The framework enables non-expert users to specify high-level semantic monitoring objectives in natural language, which are then grounded into coordinated navigation, perception, and interaction behaviors for autonomous long-horizon operation in greenhouse environments, rather than relying on manually engineered task specifications or low-level control design. Building on prior work that employs LLMs and VLMs for task planning, our approach integrates multiple sources of information—such as camera observations and top-down maps—processed in different ways to support decision making. Rather than supplying all this data to the VLM at every step, we encourage the model to query input sources selectively and efficiently. This design allows the robot to combine heterogeneous information with available navigation and active vision primitives to complete instructions successfully. Our framework supports targeted exploration and data collection driven by natural language commands, offering both flexibility and efficiency. To the best of our knowledge, this is the first work to introduce visual-language–guided crop monitoring in agricultural environments.

    

In summary, the main contributions of this paper are:
\begin{itemize}
    \item A novel, modular framework for task planning in horticultural environments that interleaves input-source queries with action-primitive callbacks.
    \item A benchmark for both short and long horizon tasks for crop monitoring in horticultural scenarios, including a lightweight simulation environment.
\end{itemize}

\section{Related works}
\label{sec:relatedworks}
\begin{table*}[htbp]
\centering
\caption{\textbf{Comparison of Methods Across Inputs, Embodiment, Deployment Setting, and Capabilities.} \\ The table below provides a comparison between our method and other approaches discussed in the literature review. While all methods contain language as an input to the model, only our method, TopV-Nav, and CoNVOI  integrate language, a map, and visual inputs to the model. Moreover, only our method and AgriVLN are deployed on a mobile manipulator in a real agricultural setting, demonstrating its applicability to unstructured, outdoor domains. Furthermore, our approach is the only one that interleaves action primitives with informational queries, enabling closed-loop, step-by-step execution of high-level semantic tasks.}
\resizebox{\textwidth}{!}{
\begin{tabular}{l@{\hskip 4mm} c c c c c c c c}
\toprule
Method & Inputs & & & Embodiment & Deployment & Capabilities & & 
\\
\toprule
& Language & Map & Vision & Mobile Agent & Agricultural Setting & Action Primitives & Informational Queries & Zero-Shot
\\
\midrule
SayCan        & $\checkmark$ & $\times$ & $\checkmark$ & $\checkmark$ & $\times$ & $\checkmark$ & $\times$ & $\checkmark$ \\
MapGPT        & $\checkmark$ & $\checkmark$ & $\times$ & $\checkmark$ & $\times$ & $\checkmark$ & $\times$ & $\checkmark$ \\
AgriVLN       & $\checkmark$ & $\times$ & $\checkmark$ & $\checkmark$ & $\checkmark$ & $\checkmark$ & $\times$ & $\checkmark$ \\
AgroGPT       & $\checkmark$ & $\times$ & $\checkmark$ & $\times$ & $\checkmark$ & $\times$ & $\times$ & $\times$ \\
TopV-Nav      & $\checkmark$ & $\checkmark$ & $\checkmark$ & $\checkmark$ & $\times$ & $\checkmark$ & $\times$ & $\checkmark$ \\
RT2           & $\checkmark$ & $\times$ & $\checkmark$ & $\checkmark$ & $\times$ & $\times$ & $\times$ & $\times$ \\
OpenVLA       & $\checkmark$ & $\times$ & $\checkmark$ & $\checkmark$ & $\times$ & $\times$ & $\times$ & $\times$ \\
$\pi_0$       & $\checkmark$ & $\times$ & $\checkmark$ & $\checkmark$ & $\times$ & $\times$ & $\times$ & $\times$ \\
NaVILA        & $\checkmark$ & $\times$ & $\checkmark$ & $\checkmark$ & $\times$ & $\checkmark$ & $\times$ & $\checkmark$ \\
Mobility VLA  & $\checkmark$ & $\times$ & $\checkmark$ & $\checkmark$ & $\times$ & $\times$ & $\times$ & $\checkmark$ \\
Behav         & $\checkmark$ & $\times$ & $\checkmark$ & $\checkmark$ & $\times$ & $\times$ & $\times$ & $\checkmark$ \\
ZeST          & $\checkmark$ & $\times$ & $\checkmark$ & $\checkmark$ & $\times$ & $\times$ & $\times$ & $\checkmark$ \\
CoNVOI        & $\checkmark$ & $\checkmark$ & $\checkmark$ & $\checkmark$ & $\times$ & $\times$ & $\times$ & $\checkmark$ \\
\midrule
\textbf{Our Method} & $\checkmark$ & $\checkmark$ & $\checkmark$ & $\checkmark$ & $\checkmark$ & $\checkmark$ & $\checkmark$ & $\checkmark$ \\
\bottomrule
\end{tabular}
}
\label{tab:method_comparison}
\end{table*}

Task planning for robotics has been extensively studied long before the rise of large language and vision-language models. Classical approaches relied on symbolic reasoning systems such as STRIPS \cite{fikes1971strips} and Hierarchical Task Networks (HTNs) \cite{kaelbling2011hierarchical, wolfe2010combined}. While these methods provide strong guarantees in structured settings, they are limited by the need for accurate, handcrafted domain models, making them unfeasible in unstructured or partially observable environments such as agricultural settings. Learning-based methods, including reinforcement learning and imitation learning, have also been explored to couple perception and planning \cite{zhang2022task}. However, they typically require large amounts of task-specific training data and generalize poorly to novel instructions or unseen environments.

More recently, large foundation models have shown impressive capabilities for navigation and manipulation tasks in a zero-shot manner, that is, without any finetuning. Their general knowledge enables them to extract relationships between objects, places, and actions, which is fundamental for planning and decision making. They have also enabled instruction following from natural language commands. We focus this review on such models, and more specifically on the use of LLMs and VLMs for task planning in mobile-manipulation agents.
A comparison of our method with other approaches is presented in Table \ref{tab:method_comparison}.

\textbf{LLMs and VLMs in General Domains}

LLMs have demonstrated task planning capabilities since the early days of these models. For instance, \cite{huang2022language} showed that LLMs can decompose language instructions into mid-level plans for mobile-manipulation tasks in indoor environments, using only a few demonstration examples of related tasks and without additional training. Several approaches have emerged to improve the performance of these models, including the use of pre-trained value functions to weight the skills proposed by LLMs for better grounding in the real world \cite{ahn2022can}. LLMs can also take feedback from different sources (e.g., scene description models, object detectors, or humans in the loop) to improve problem-solving tasks for embodied agents \cite{huang2022inner}. Additionally, the coding capabilities of these models have been leveraged to generate language model programs (LMPs) that manage action primitives to solve complex and long-horizon tasks \cite{liang2022code}. 

With the advent of VLMs, models no longer require explicit scene descriptions as input, since they can directly process images, overcoming a key limitation of LLMs. Simple approaches take images as input with potential actions superimposed on them (e.g., polar actions in \cite{goetting2024end}). The models then either assign scores to these actions or directly choose one for execution \cite{nasiriany2024pivot, goetting2024end}, simplifying the classical perception–planning–control pipeline. VLMs have also been leveraged to extract various forms of information from camera observations, including object descriptions \cite{wu2024voronav}, success/failure detection \cite{zhi2024closed}, action-conditioned relationships between objects (e.g., opening a drawer reveals a toy) \cite{jiang2024roboexp}, and traversability estimation \cite{elnoor2024robot}. 

More complex methods incorporate maps to support decision making. Different types of maps have been proposed. Visual language maps, typically dense representations with visual–language features (e.g., CLIP embeddings \cite{radford2021learning}), enable object position retrieval, where an LLM can parse the instructions into object queries \cite{huang2022visual}. Similarly, topological maps and 3D scene graphs can store semantic features in their nodes more efficiently, making them suitable for larger environments \cite{shah2023lm, werby2024hierarchical}. While these maps allow objects to be retrieved in natural language, they provide limited spatial context to the LLM.

To address the lack of spatial awareness in previous works, \cite{chen2024mapgpt} proposed MapGPT, where a topological map encoded in linguistic form is provided to the LLM. This enables the agent to understand spatial structures to some degree, but still omits essential metric information such as object shapes, sizes, and explored/unexplored regions. A closer approach to ours is TopV-Nav \cite{zhong2024topv}, which uses a top-down map containing key areas, objects labels, historical traversed regions, and frontier segments as input to a VLM. The model assigns scores to key areas related to the location of a target object. In contrast, we show that combining global maps, egocentric maps, and camera observations allows a VLM to directly execute action primitives for exploration and monitoring tasks. 

\textbf{LLMs and VLMs for Question Answering}

One of the earliest efforts to connect language models with robotic planning is SayCan \cite{ahn2022can}, which combines LLM-generated high-level instructions with affordance value functions representing the feasibility of pre-trained robotic skills. By grounding semantic reasoning in physical affordances, SayCan enabled mobile manipulators to perform over a hundred real-world kitchen tasks with impressive planning accuracy. Building on this insight, the CAPE \cite{raman2024cape} framework introduced corrective planning mechanisms, allowing an LLM to suggest recovery actions when preconditions were violated. CAPE significantly improved robustness across both simulated and real-world environments, demonstrating how retrieval and feedback loops can address weaknesses in static LLM-driven plans.

Parallel work has explored structuring robot policies to increase interpretability and resilience. For instance, BETR-XP-LLM \cite{styrud2024automatic} dynamically expands behavior trees through LLM-guided reasoning, enabling robots to generate transparent and verifiable plans that adapt to execution challenges. By representing actions as structured tree nodes, this line of work connects the flexibility of language models with the reliability of symbolic planning, offering a more trustworthy substrate for embodied decision-making.

Recent studies have extended these ideas to embodied question answering (EQA) and long-horizon reasoning. OpenEQA \cite{majumdar2024openeqa} formalized EQA benchmarks to evaluate foundation models in robotic environments, positioning language-guided visual reasoning as a critical capability. Building on this benchmark, \cite{ginting2025enter} introduced memory-augmented reasoning that allows robots to recall and plan across extended temporal horizons. Similarly, \cite{lan2025experience} proposed self-generated memory as a way to ground VLM predictions in past experience, improving fidelity in sequential tasks. Collectively, these works shift the focus from single-shot responses toward temporally coherent reasoning, which is essential for embodied applications.

Another frontier in this space integrates visual grounding and closed-loop control. \cite{hu2023look} provided early evidence that VLMs can generate high-level robotic plans from visual input, establishing the viability of multimodal foundation models for action planning. In \cite{zhi2024closed}, the  framework proposed extended this approach by embedding feedback mechanisms into VLM-driven manipulation, allowing robots not only to propose actions but also to verify success and adjust dynamically. These studies demonstrate the growing role of vision-language models in bridging perception and planning through feedback, complementing the affordance-based and memory-based approaches described above.

Yet despite these advances, little work explores how such embodied reasoning can be applied in agricultural contexts, where the need to integrate perception, planning, and execution is equally critical but underexplored.

\textbf{LLMs and VLMs in Agricultural Environments}

The application of LLMs and VLMs in agriculture has primarily focused on disease and pest detection, rather than full robotic autonomy \cite{haghighat2025multimodal}. Several recent works have aimed to adapt foundation models for agricultural domains. For instance, AgriCLIP adapts CLIP via domain-specialized cross-model alignment, improving performance on tasks involving crops and livestock by leveraging agricultural-specific image-text pairs \cite{nawaz2024agriclip}. Similarly, AgroGPT demonstrates an efficient agricultural vision-language model with expert tuning, allowing the model to handle diverse visual inspection tasks with high accuracy \cite{awais2025agrogpt}. Other studies have explored VLMs for crop disease diagnosis: a visual large language model for wheat disease detection in the wild \cite{zhang2024visual}, SCOLD for leaf disease identification \cite{quoc2025vision}, and few-shot image classification of crop diseases using vision-language models \cite{zhou2024few}. These approaches indicate the potential of VLMs to generalize across multiple crops and disease types with limited labeled data.

To systematically evaluate these models, agricultural benchmarks such as AgroBench provide a broad set of 682 disease categories across multiple crops, enabling consistent comparison of model performance in zero-shot and few-shot settings \cite{shinoda2025agrobench}. SCOLD further contributes with 186,000 image-text pairs for fine-grained disease identification \cite{quoc2025vision}. AgriVLM, a separate framework, has also achieved over 90\% accuracy in identifying disease and growth stages, demonstrating the promise of vision-language reasoning in practical agricultural applications \cite{yu2024framework}.

While most prior work focuses on classification and recognition, only a limited number of studies have explored integrating these models into robotic pipelines, particularly in agricultural settings. AgriVLN enables visual-language navigation in agricultural scenes, decomposing language instructions into subtasks and outputting low-level actions to navigate from start to destination \cite{zhao2025agrivln}. \cite{zuzuarregui2025leveraging} propose mission planning in agricultural environments using an LLM with textual farm layouts, but find spatial reasoning is limited when key spatial features are provided only in text. In contrast, our approach allows a VLM to query multiple sources information actively, improving crop monitoring efficiency while integrating perception, planning, and action in real-time.

\textbf{Vision-Language-Action and Vision-Language-Navigation Models}

Vision-language-action (VLA) models have emerged as a promising end-to-end approach for navigation and manipulation tasks. Pioneering works such as RT-1 \cite{brohan2022rt} take sequences of images and text descriptions as input and directly output joint manipulation commands and low-level navigation actions to perform mobile manipulation tasks in indoor environments. More recent works like RT-2 \cite{zitkovich2023rt}, OpenVLA \cite{kim2024openvla} and $\pi_0$ \cite{black2024pi_0} leverage pre-trained vision-language models as the primary backbone for VLA systems, which are then fine-tuned on multi-modal and robotic data to generate action tokens.

While these methods demonstrate impressive generalization across tasks and embodiments, their effectiveness is tied to the availability of large-scale training datasets, including both synthetic and real-world demonstrations. In practice, this requires extensive collections of both real-world and synthetic demonstrations tailored to specific robot embodiments, sensor configurations, and environments. Acquiring such data necessitates additional fine-tuning, which introduces susceptibility to out-of-distribution challenges which is a concern in agricultural settings, where plants undergo continuous morphological and chromatic variation as they mature and fruit ripens. Zero-shot models, trained on substantially larger and more diverse corpora, offer a potential pathway to mitigate these generalization limitations.

Recent iterations have sought to improve specialized capabilities; for instance, NaVILA \cite{cheng2024navila} utilizes a hierarchical structure to bridge high-level reasoning with mid-level language actions for legged locomotion, while Mobility VLA \cite{chiang2024mobility} incorporates long-context reasoning over topological graphs to improve navigation robustness. These methods address the abstraction in reasoning but are mostly dedicated to navigation tasks.

In contrast, our modular approach leverages the common-sense reasoning capabilities of VLMs for high-level planning, while relying on classical low-level planners and controllers to ensure safe navigation and viewpoint execution along collision-free trajectories. Compared to VLA models, our approach does not require large-scale datasets for fine-tuning and is able to perform crop monitoring tasks in a zero-shot setting.

Parallel to VLA research, vision-language-navigation (VLN) methods have explored the use of language-conditioned reasoning for navigation. Several recent works address this problem, including Behav \cite{weerakoon2025behav}, ZeST \cite{gummadi2025zest}, CoNVOI \cite{sathyamoorthy2024convoi} and others \cite{elnoor2025vlm, song2024vlm} . Most approaches rely on LLMs or VLMs to decompose natural language instructions into semantic regions and estimate traversability scores for each region. Combined with open-vocabulary segmentation, this enables the construction of traversability maps, which are then used by classical path planners.

Our approach similarly treats navigation as a modular component but differs in two key aspects. First, navigation is integrated into the broader task planning problem for crop monitoring. Second, we leverage occupancy grid maps instead of learning-based traversability maps, resulting in lower computational overhead. This enables a more holistic system capable of operating reliably in real-world agricultural settings.

\section{Methods}
\label{sec:methods}
\begin{figure*}[htp]
    \centering
    \includegraphics[width=1\textwidth]{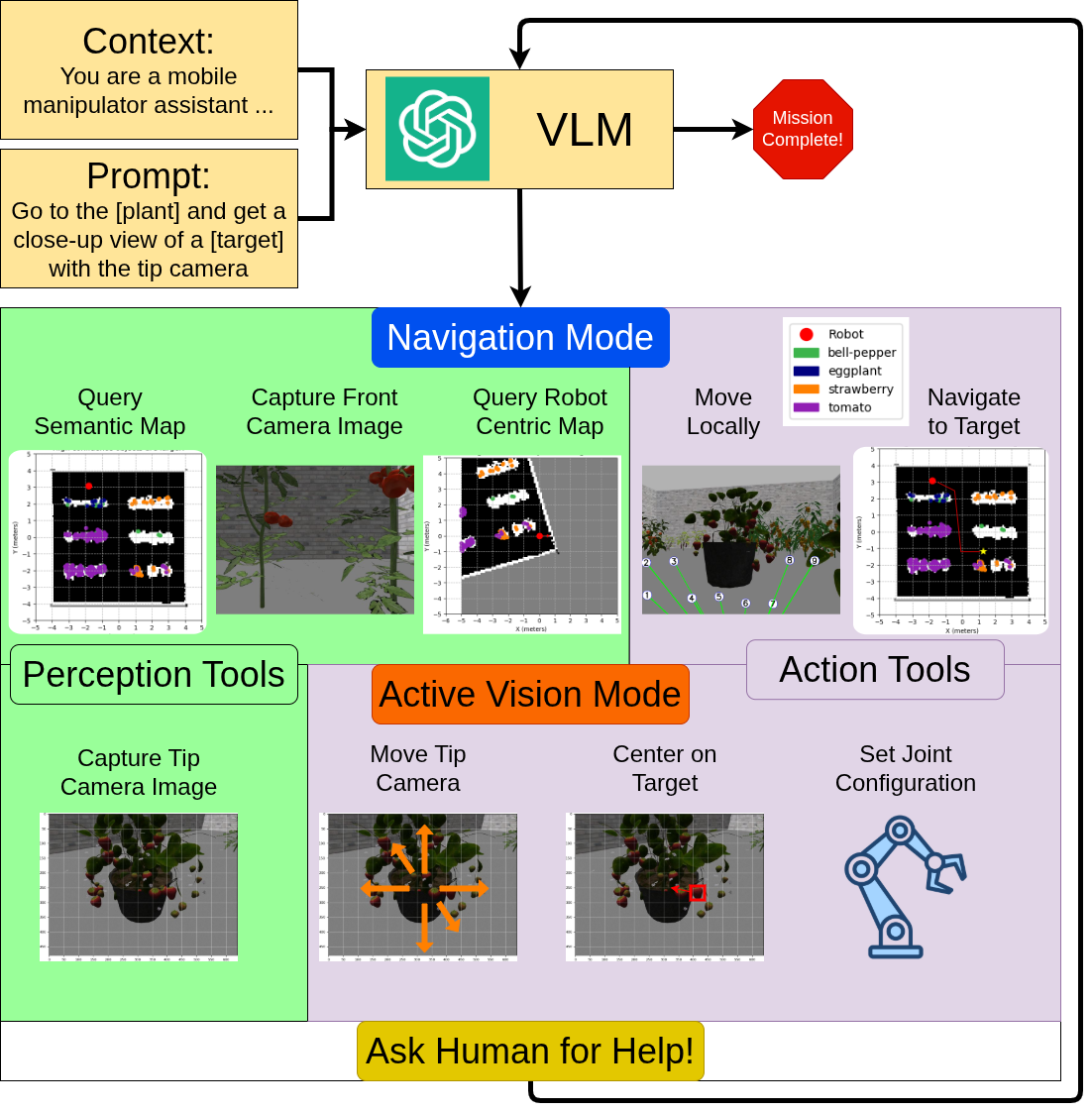}
    \caption{\textbf{Overall System Pipeline} Given context and the assigned task, the agricultural agent selects from a library of perception and action primitives to achieve the task. The agent can switch between active vision} and navigation modes. Once the task is complete or the agent requires help, a human can be prompted to intervene.
    \label{fig:AgAgent-Pipeline}
\end{figure*}

\begin{figure*}[htp]
    \centering
    \includegraphics[width=1\textwidth]{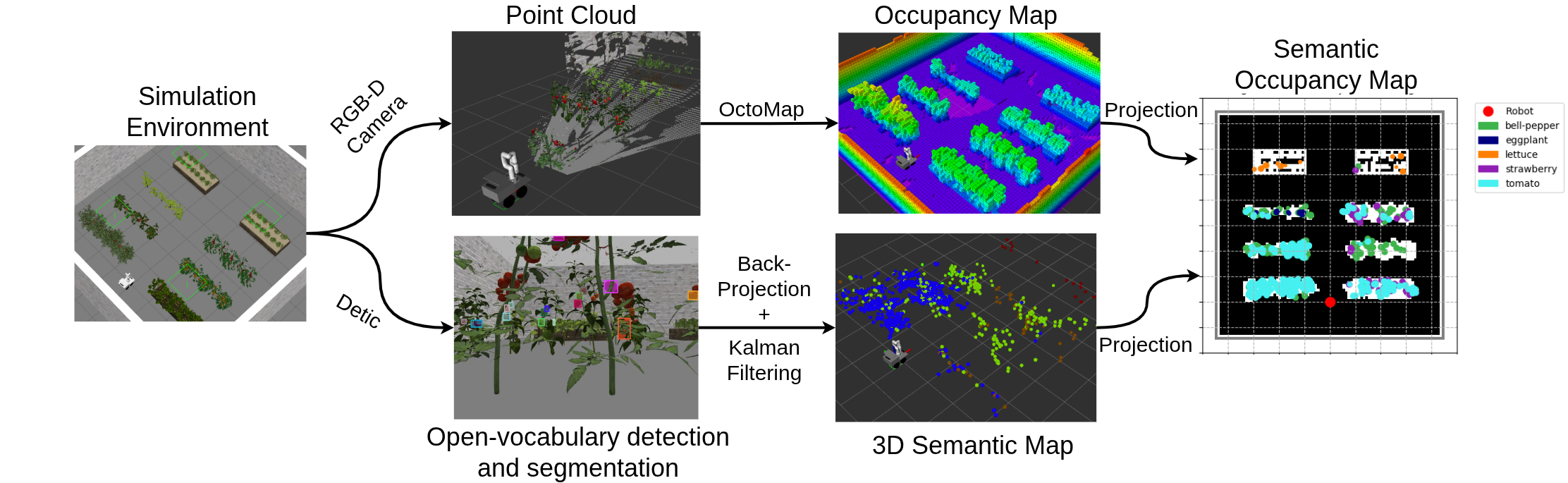}
    \caption{Semantic mapping pipeline. The RGBD data is used to generate an occupancy map of the environment. Detic is used for open vocabulary object detection and segmentation. The output is backprojected and filtered with kalman filtering.  Both branches are projected to create a  semantic occupancy map. }
    \label{fig:Semantic Map}
\end{figure*}


In this work, we aim to perform crop monitoring tasks in horticultural environments using a mobile manipulator controlled by a VLM. Specifically, the robot starts from an arbitrary position in the environment and is given a crop monitoring task defined by one or more plant targets (e.g., tomato or bell pepper plants) and one or more inspection targets (e.g., a tomato cluster or a stem). The objective is for the VLM to leverage a set of perception and action tools, as well as a pre-built semantic occupancy map to efficiently accomplish the assigned task.

Figure \ref{fig:AgAgent-Pipeline} presents an overview of our system. The robotic platform consists of a mobile manipulator equipped with two RGBD cameras: one fixed to the front of the robot base and another mounted on the manipulator's end-effector. Our approach focuses on providing the VLM with a diverse set of perception and action tools that enable decision-making for both global and local navigation, as well as precise target inspection using the end-effector camera. A key component of our system is a semantic occupancy map, which encodes information about both scene occupancy and object locations. This map serves two main purposes: it supports the classical navigation stack to ensure safe and efficient path planning, and it provides spatial context for high-level decision-making by the VLM.

\subsection{Perception}
\label{sec:perception}
\textbf{Open-Vocabulary Semantic Occupancy Map}. Inspired by prior work on semantic mapping that leverages visual-language features (e.g., CLIP \cite{radford2021learning}) to create open-vocabulary maps \cite{huang2022visual}, we build a semantic map of sparse objects modeled as spheres and semantic feature vectors. Our mapping approach is presented in Fig. \ref{fig:Semantic Map}. We use Detic\cite{zhou2022detecting} for open-vocabulary detection and segmentation. This allows us to construct semantic maps with a customizable set of categories without fine-tuning on specific crop types. Each detected object is tracked over time using a Kalman filter, which fuses multiple observations to update its position, size, and confidence. This combination of an efficient detector and compact object representation allows our mapping pipeline to operate in real time, a critical property for robotic applications.

In parallel, we maintain an OctoMap \cite{hornung2013octomap} built from RGBD observations, encoding free, occupied, and unknown regions. By projecting both the 3D semantic objects and the occupancy map into a top-down view, we generate a 2D semantic occupancy map that integrates occupancy information, object locations, and semantic confidence. This map is provided to the VLM for decision-making, supporting exploration and object-goal navigation tasks.

\textbf{Enriched Camera Image Observations}. To address the spatial reasoning limitations of current VLMs, we preprocess images from both the base and end-effector cameras. Following prior works \cite{goetting2024end}, we project polar action commands from the robot coordinate frame onto the base camera image plane. This enables the VLM to reason directly in the image space and navigate toward visible target points. For the tip camera, we superimpose a 2D grid onto the image, allowing the VLM to more precisely localize objects in image coordinates. These 2D locations are then projected into 3D coordinates using the corresponding depth image and camera intrinsics. This preprocessing step allows the VLM to plan both navigation and active vision actions in the image domain, which are then executed by the robot's low-level controllers.

\subsection{Planning}
\textbf{Task Planner}. We employ a VLM agent to orchestrate a set of perception and action primitives in order to accomplish a task. The model receives three main inputs: a context description, descriptions of available perception and action primitives, and the task instruction in natural language. 

Unlike prior approaches that generate a full program of primitive actions (i.e., a language model program, or LMP \cite{liang2022code}), our VLM agent operates iteratively: at each time step, it outputs the next tool or task to execute, receives feedback from the execution, and performs online replanning. This design is crucial because horticultural environments are partially observable. For instance, a fruit of interest might be occluded by leaves or out of the current field of view, making predefined action sequences unreliable without feedback. We also maintain a history of tool calls and outcomes, allowing the agent to reason over past experiences. However, for efficiency and relevance, we only include the most recent image observation in the context window.

\textbf{Context description}. This includes information about the robot configuration, available data sources (e.g., map, RGBD images), and a set of behavioral rules promoting safety, motion efficiency, high-quality image capture, failure recovery, and proper stopping conditions.

\textbf{Tool descriptions}. For each perception or action tool, we specify its purpose, input arguments, and expected outputs. We also prompt the VLM to justify each tool selection with a brief reasoning statement to enhance interpretability and traceability of its decisions.

\textbf{Task specification}. Tasks are provided in natural language and typically involve one or more navigation targets (e.g., specific plants) and inspection targets (e.g., fruits, stems or leaves).

\textbf{Navigation and Active Vision} Primitives. We implement a set of tools for global navigation (e.g., navigate-to-map-point), local navigation (e.g., rotate-and-move-forward), and active vision (e.g., move-tip-camera-left). 

While the semantic map allows the VLM to identify the approximate location of objects, relying solely on it is insufficient because the map may contain false positives or inaccurate object positions due to perception noise. Therefore, we also provide robot-centric representations, enabling more accurate localization and short-range navigation adjustments (e.g., fine-tuning orientation or moving forward). Alternatively, the base camera image with superimposed polar actions described in the previous section can be used for local motion planning.

For active vision, we include primitive motions for the tip camera along the six Cartesian directions (left, right, up, down, forward, and backward). We also design a centering tool that automatically aligns an object of interest with the camera's optical center providing a more efficient and flexible alternative to fixed-step movement commands.

The prompts, context descriptions, and tool descriptions are provided in the supplementary material.

\subsection{Control}
For low-level motion planning and execution, we employ the ROS Navigation Stack \cite{ros-move_base} and the MoveIt \cite{coleman2014reducing} frameworks. The navigation stack provides global path planning, enabling the mobile platform to generate collision-free trajectories through the occupancy grid map. In parallel, MoveIt supports kinematic modeling, motion planning, and collision checking for the manipulator, allowing safe and efficient arm movements within the same mapped environment. By leveraging a shared environmental representation, the subsystems coordinate base navigation and arm motion to support active vision tasks throughout the simulation environment.

\section{Experiments}
\begin{figure*}[htp]
    \centering
    \includegraphics[width=1\textwidth]{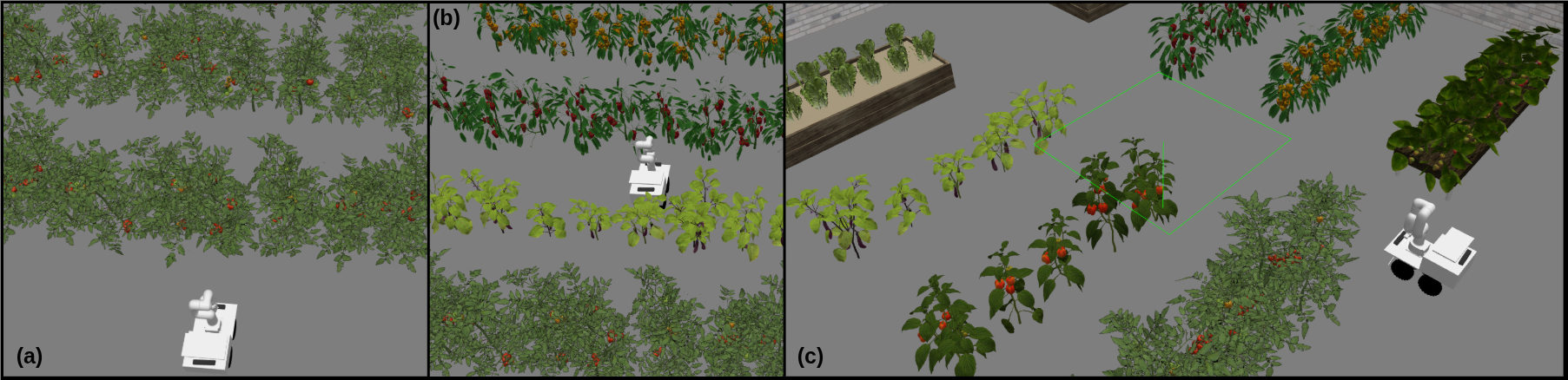}
    \caption{\textbf{Simulation Environments} Three gazebo simulation environments are used to evaluate the system. (a) A monoculture environment filled with various ripe and unripe tomatoes. (b) A polyculture environment filled with tomatoes, orange peppers, red peppers, and eggplants. (c) A polyculture environments filled with tomatoes, green peppers, red peppers, and eggplants as well as lettuce and strawberries on raised tables.}
    \label{fig:Simulation-Environments}
\end{figure*}

\subsection{Simulation Environment}
We evaluated our system in three Gazebo simulation environments modeled after realistic greenhouse layouts. Sample viewpoints from each scene are shown in Fig. \ref{fig:Simulation-Environments}. In all scenes, crops are arranged in rows spaced 1.5 m apart center-to-center.
The first and simplest scene is a tomato monoculture. Tomatoes are a high-value crop commonly grown this way \cite{kaiser2012high, waterer2003yields}. Tomato plants with varying geometry and fruits at different ripeness stages were randomly placed throughout four rows.
The second scene increases complexity by introducing a polyculture: tomatoes, eggplants, orange bell peppers, and red bell peppers. Each crop type occupies a single row, resulting in four distinct rows and greater visual variety for the VLM.
The third and most complex scene includes all crops from the second scene and adds lettuce and strawberries. These additional crops are placed in raised planters, reflecting common agricultural practice and contributing more geometric variation. The six crop varieties each occupy their own row, with an additional row of tomatoes and lettuce. The scene is arranged as two columns of four rows, with a central aisle to improve access to the plants.

\subsection{Metrics}
We report both the \textit{success rate} and the \textit{task completion rate} as measures of system effectiveness. The success rate is defined at the episode level and indicates whether the robot successfully completes the overall task objective within a trial. In contrast, the task completion rate is defined at the subgoal level and measures the proportion of intermediate objectives (e.g., navigation targets or inspection viewpoints) that are successfully achieved during task execution. We also report the number of tool calls as an indicator of efficiency, and finally, the success rate weighted by shortest path length as an indicator of navigation success and efficiency.

To compute the success rate and task completion rate, we divide each task into multiple subgoals, including both navigation and active vision subgoals. For example, the task \textit{``navigate to three tomato plants and take two pictures from each — one of a green tomato and one of a ripe tomato''} is split into three navigation subgoals and six active vision subgoals. An episode is considered successful if all of its subgoals are successfully completed. The success rate $SR$ is computed as:
\[
SR = \frac{1}{N}\sum_{i=1}^{N}S_i
\]
where $S_i \in \{0,1\}$ is the success indicator for episode $i$, and $N$ is the total number of episodes.

The completeness score $C_i$ for episode $i$ is computed as the ratio between the number of successfully completed subgoals and the total number of subgoals in that episode. We then compute the average task completion rate $TCR$ as:
\[
TCR = \frac{1}{N}\sum_{i=1}^{N}C_i
\]

We report the number of tool calls as the total number of perception and action primitives invoked to complete a task. This metric is averaged across the successful episodes only, since partially completed tasks can lead to an underestimation of the actual number of required tool calls.

We compute the success rate weighted by shortest path length $SPL$ following \cite{batra2020objectnav}:
\[
SPL = \frac{1}{N}\sum_{i=1}^{N}S_i\frac{l_i}{\max(l_i, p_i)}
\]
where $p_i$ is the distance traveled by the agent. Unlike \cite{batra2020objectnav}, where $l_i$ represents the shortest path length, we define $l_i$ as the distance traveled by the human expert. This is due to the fact that finding the true optimal path requires identifying the optimal sequence of plant targets to visit, which is cumbersome in our environments containing multiple crop instances distributed across different rows. Despite this assumption, our $SPL$ remains a reliable indicator of navigation success and efficiency relative to human experts, as it penalizes long or inefficient trajectories while rewarding both successful and efficient navigation behavior.

\subsection{Benchmark Description}

\textbf{Task categories}. We split the tasks into four types based on complexity: (1) Single Plant, Single Target, (2) Single Plant, Multiple Targets, (3) Multiple Plants, Single Target, and (4) Multiple Plants, Multiple Targets. Each task prompt requires the robot to navigate to a specified plant and monitor one or more target parts of that plant. We randomize the plant type, the target plant parts (leaves, stems, fruits), the number of plants (one to three), the number of target parts (one to three), characteristics of the target part (e.g., ripe tomato, red bell pepper), and the initial robot pose within the environment. To increase linguistic diversity, we use an LLM to paraphrase the natural-language prompts. We run 20 prompts per environment for tasks involving a single plant, and 10 prompts per environment for tasks involving multiple plants. In total, the evaluation consists of 198 tasks, divided into three groups and executed by three human experts and the VLM agent.

\textbf{Stopping condition}. Unlike prior work that uses a dedicated VLM for success evaluation, in our setup human experts supervise all tasks and manually stop execution when the task is successfully completed, unless the VLM explicitly reports completion. Because the VLM can get stuck in infinite loops during execution, we define an early-stopping condition that instructs the VLM to skip the current subgoal and continue to the next one. Specifically, if the VLM fails to accomplish a subgoal after nine tool calls since the previous subgoal (a threshold derived from human performance statistics), we inform the VLM to abort that subgoal.

\textbf{VLM agent}. We use GPT-4.1 as the VLM agent via the OpenAI API with tool-calling capabilities.

\textbf{Evaluation Under Noiseless Conditions}. In this study, we evaluate the model in a zero-shot setting and use the robot-centric map as a tool for local navigation. Additionally, we provide a pre-built occupancy map with ground-truth object locations to analyze the VLM’s decision-making performance under ideal conditions.

\textbf{Evaluation Under Noisy and Variant Conditions.} We evaluate the performance of our method under noisy conditions, specifically using a semantic map that contains false positives introduced by the object detector. This scenario is particularly challenging for a VLM agent, which relies heavily on the estimated locations of objects. When the robot navigates toward a target object that is actually a false positive, the VLM must 1) verify whether the object is an actual false positive in the map, 2) perform local exploration to search for the correct object nearby, and 3) if unsuccessful, navigate to an alternative map location for the same object.

We conduct seven trials per task category using the noisy map produced by our mapping framework. For each task, we evaluate three agents: a human operator, a VLM-zero-shot agent, and a VLM-few-shot agent. In contrast to the zero-shot setting, the VLM-few-shot agent receives example reasoning demonstrations in addition to the context description. These demonstrations illustrate situations such as reaching an incorrect navigation target, failing to execute a tool, or detecting a plant in front of the robot but outside the manipulator's reachable workspace. To reduce token usage, the demonstrations include only natural-language scene descriptions and reasoning steps, without images. The full demonstrations are provided in the supplementary material.

We also consider two modes for local navigation: one using a robot-centric map, and another using the front camera augmented with potential actions. Both modes provide guidance for forward motion and in-place rotations, as described in Section \ref{sec:perception}. For this study, we use the most complex polyculture environment, which includes more crop types and therefore produces noisier detections. A single precomputed noisy map is used for all runs to ensure fair comparison.

As in earlier experiments, we stop the agent once the task is successfully completed. Additionally, if the agent fails to reach a navigation target after nine tool calls, we halt execution and instruct the agent to skip that target.


\section{Results and Discussion}
\subsection{Results under Noiseless Conditions}

\begin{table*}[htp!]
\centering
\caption{\textbf{Main Results Across All Environments}}
\resizebox{\textwidth}{!}{
    \begin{tabular}{l@{\hskip 4mm} c c c c c c c c c}
    \toprule
Task Category & Operator & Number of Trials & Success Rate (\%) $\uparrow$ & Number of Tool Calls $\downarrow$ & Task Completion Rate (\%) $\uparrow$ & SPL $\uparrow$\\
    \midrule
    \multirow{2}{*}{Single Plant - Single Target} & Human & 62 & 91.94 & 9.53$\pm$3.15 & 96.27$\pm$15.36 & 90.32\\
    & Zero-Shot VLM & 60 & 86.67 & 11.33$\pm$3.04 & 96.72$\pm$11.24 & 60.84\\
    \midrule
    \multirow{2}{*}{Single Plant - Multiple Target} & Human & 60 & 91.67 & 14.04$\pm$5.02 & 97.67$\pm$8.44 & 91.67\\
    & Zero-Shot VLM & 57 & 68.42 & 18.38$\pm$5.28 & 88.33$\pm$20.97 & 47.64\\
    \midrule
    \multirow{2}{*}{Multiple Plant - Single Target} & Human & 43 & 93.02 & 22.95$\pm$2.58 & 98.58$\pm$5.73 & 93.02\\
    & Zero-Shot VLM  & 38 & 42.11 & 33.44$\pm$7.79 & 85.05$\pm$15.12 & 28.63\\
    \midrule
    \multirow{2}{*}{Multiple Plant - Multiple Target} & Human & 33 & 84.85 & 33.11$\pm$6.78 & 90.06$\pm$17.16 & 84.85\\
    & Zero-Shot VLM & 32 & 9.38 & 43.33$\pm$9.29 & 76.19$\pm$17.04 & 5.87\\
    \midrule
    \multirow{2}{*}{Average Over All Categories} & Human & 198 & 90.37 & 19.91$\pm$4.38 & 97.14$\pm$11.67 & 89.97\\
    & Zero-Shot VLM & 187 & 51.64 & 26.62$\pm$6.35 & 86.57$\pm$16.09 & 35.74\\

    \bottomrule
    \end{tabular}
    }
\label{tab:main.results}
\end{table*}

\begin{table*}[htp!]
\centering
\caption{\textbf{Main Results Complex Polyculture Environment}}
\resizebox{\textwidth}{!}{%
    \begin{tabular}{l@{\hskip 4mm} c c c c c c c c c}
    \toprule
Task Category & Operator & Number of Trials & Success Rate (\%) $\uparrow$ & Number of Tool Calls $\downarrow$ & Task Completion Rate (\%) $\uparrow$ & SPL $\uparrow$\\
    \midrule
    \multirow{2}{*}{Single Plant - Single Target} & Human & 21 & 100.0 & 8.14$\pm$1.21 & 100.0$\pm$0.0 & 100.0\\
    & Zero-Shot VLM & 21 & 100.0 & 10.24$\pm$2.72 & 100.0$\pm$0.0 & 68.44\\
    \midrule
    \multirow{2}{*}{Single Plant - Multiple Target} & Human & 21 & 100.0 & 11.81$\pm$2.52 & 100.0$\pm$0.0 & 100.0\\
    & Zero-Shot VLM & 21 & 80.95 & 16.94$\pm$3.04 & 92.86$\pm$15.70 & 52.97\\
    \midrule
    \multirow{2}{*}{Multiple Plant - Single Target} & Human & 19 & 100.0 & 22.74$\pm$1.37 & 100.0$\pm$0.0 & 100.0\\
    & Zero-Shot VLM  & 19 & 31.58 & 38.33$\pm$10.81 & 81.58$\pm$15.08 & 24.65\\
    \midrule
    \multirow{2}{*}{Multiple Plant - Multiple Target} & Human & 11 & 100.0 & 32.45$\pm$6.53 & 100.0$\pm$0.0 & 100.0\\
    & Zero-Shot VLM & 11 & 27.27 & 43.33$\pm$9.29 & 83.55$\pm$15.81 & 17.07\\

    \bottomrule
    \end{tabular}}
\label{tab:complex.polyculture}
\end{table*}

\begin{table*}[htp!]
\centering
\caption{\textbf{Main Results Simple Polyculture Environment}}
\resizebox{\textwidth}{!}{%
    \begin{tabular}{l@{\hskip 4mm} c c c c c c c c c}
    \toprule
Task Category & Operator & Number of Trials & Success Rate (\%) $\uparrow$ & Number of Tool Calls $\downarrow$ & Task Completion Rate (\%) $\uparrow$ & SPL $\uparrow$\\
    \midrule
    \multirow{2}{*}{Single Plant - Single Target} 
        & Human & 21 & 85.71 & 12.28$\pm$4.08 & 93.76$\pm$21.47 & 80.95\\
        & Zero-Shot VLM & 19 & 73.68 & 14.07$\pm$2.25 & 97.53$\pm$5.67 & 57.44\\
    \midrule
    \multirow{2}{*}{Single Plant - Multiple Target} 
        & Human & 19 & 94.74 & 18.00$\pm$5.13 & 99.21$\pm$3.35 & 94.74\\
        & Zero-Shot VLM & 17 & 70.59 & 20.08$\pm$8.28 & 88.82$\pm$25.81 & 55.54\\
    \midrule
    \multirow{2}{*}{Multiple Plant - Single Target} 
        & Human & 14 & 92.86 & 24.00$\pm$3.74 & 99.21$\pm$2.83 & 92.86\\
        & Zero-Shot VLM & 9 & 33.33 & 29.67$\pm$1.89 & 81.44$\pm$16.49 & 26.18\\
    \midrule
    \multirow{2}{*}{Multiple Plant - Multiple Target} 
        & Human & 13 & 84.62 & 29.64$\pm$5.07 & 91.85$\pm$26.56 & 84.62\\
        & Zero-Shot VLM & 11 & 0.00 & NA & 66.73$\pm$16.33 & 0.00\\
    \bottomrule
    \end{tabular}}
\label{tab:simple.polyculture}
\end{table*}

\begin{table*}[htp!]
\centering
\caption{\textbf{Main Results Monoculture Environment}}
\resizebox{\textwidth}{!}{%
    \begin{tabular}{l@{\hskip 4mm} c c c c c c c c c}
    \toprule
Task Category & Operator & Number of Trials & Success Rate (\%) $\uparrow$ & Number of Tool Calls $\downarrow$ & Task Completion Rate (\%) $\uparrow$ & SPL $\uparrow$\\
    \midrule
    \multirow{2}{*}{Single Plant - Single Target} 
        & Human & 20 & 90.00 & 8.39$\pm$1.42 & 95.00$\pm$15.00 & 90.00\\
        & Zero-Shot VLM & 20 & 85.00 & 10.41$\pm$2.52 & 92.50$\pm$17.85 & 56.08\\
    \midrule
    \multirow{2}{*}{Single Plant - Multiple Target} 
        & Human & 20 & 80.00 & 12.50$\pm$4.69 & 93.75$\pm$13.40 & 80.00\\
        & Zero-Shot VLM & 19 & 52.63 & 18.80$\pm$1.94 & 82.89$\pm$19.95 & 34.68\\
    \midrule
    \multirow{2}{*}{Multiple Plant - Single Target} 
        & Human & 10 & 80.00 & 21.75$\pm$1.64 & 95.00$\pm$10.62 & 80.00\\
        & Zero-Shot VLM & 10 & 70.00 & 30.86$\pm$1.96 & 94.90$\pm$7.79 & 38.39\\
    \midrule
    \multirow{2}{*}{Multiple Plant - Multiple Target} 
        & Human & 9 & 66.67 & 40.67$\pm$3.09 & 97.33$\pm$3.77 & 66.67\\
        & Zero-Shot VLM & 10 & 0.00 & NA & 78.50$\pm$14.07 & 0.00\\
    \bottomrule
    \end{tabular}}
\label{tab:monoculture.results}
\end{table*}

\textbf{Result 1: Human operators consistently outperform zero-shot VLMs across all metrics}
Human operators demonstrate consistently superior performance across all task categories and environments. Averaged over the full task set, human success rates are high for both simple and complex scenarios, including multi-target and multi-plant conditions (Last row of Table \ref{tab:main.results}). In contrast, the zero-shot VLM exhibits substantial degradation as task complexity increases. While VLM performance is adequate for simpler tasks, success rates drop to 42.11\% in multi-plant single-target problems and to 9.38\% in multi-plant multi-target scenarios (Rows 3 and 4 of Table \ref{tab:main.results}).

This gap is preserved even in visually complex environments. In the complex polyculture environment, humans maintain 100\% success across all categories (Table \ref{tab:complex.polyculture}), whereas VLM success ranges from 80.95-100\% (Rows 1 and 2 of Table \ref{tab:complex.polyculture})in single-plant tasks down to 27.27–31.58\% in multi-plant tasks (Rows 3 and 4 of Table \ref{tab:complex.polyculture}). SPL values follow the same trend, with humans consistently producing more direct and efficient trajectories. These findings highlight the robustness and adaptability of human operation compared to the brittleness of the zero-shot VLM under increasing environmental and task complexity.

Neither the human nor the VLM achieves 100\% success in the simple polyculture or the monoculture. This is due to the experimental design defining a failure when the task required finding a specific color tomato on a selected plant. If the target was a green or yellow tomato, and the plant selected only had ripe red tomatoes, then this active vision task was considered a failure. This choice was deliberately implemented to simplify the computation of metrics and avoid the replanning required in scenarios where the target was not present. Although this choice resulted in recorded "failures" that were artifacts of the task definition rather than actual performance errors, it was maintained to ensure a streamlined and consistent evaluation framework. Notably, this specific failure scenario did not occur during the human experiments in the complex polyculture environment, which consequently achieved a 100\% success rate under the defined task parameters.

\textbf{Result 2: VLM control performs reliably on simple single-plant, single-target tasks}

Despite its limitations, the zero-shot VLM performs reliably on the simplest task category involving a single plant and a single target. Across all environments, the VLM achieves 86.7\% average success—approaching the 92\% achieved by human operators (Row 1 of Table \ref{tab:main.results}). In the complex polyculture environment, the VLM even reaches 100\% success on this task, matching human performance.

These tasks represent minimal-horizon navigation problems where the target is visually salient and the required action sequence is short (see a sample episode in Fig. \ref{fig:sample_episode}). Under these conditions, the VLM generates coherent and stable plans, reflected in consistent task completion rates and modest tool call counts relative to more complex scenarios. This result demonstrates that zero-shot vision-language-grounded control is viable for simple, well-structured agricultural tasks.

\textbf{Result 3: VLMs are less efficient than Human Operators}

Across all task categories, VLM control is less efficient than human teleoperation, as reflected in tool call counts and SPL metrics. Human operators typically complete tasks using about 9-11 tool calls for simple cases and 23–33 calls for multi-plant problems. The VLM consistently exceeds these values, requiring 11–18 calls for simple tasks and 33–43 calls in complex settings.

Higher tool call counts indicate more frequent replanning, corrective actions, and uncertain decision-making. Correspondingly, the VLM’s SPL scores remain well below those of humans, even in successful trials. This pattern suggests that while the VLM can infer reasonable high-level strategies, it lacks the precision and spatial efficiency of human operators, leading to longer and more circuitous behaviors. These behavioral inefficiencies present an important limitation for real-time robotic deployment.

\textbf{Result 4: Environmental Complexity Degrades VLM Performance}

Environmental complexity has a pronounced and disproportionate effect on VLM performance. In the most challenging environment —characterized by dense foliage, occlusion, irregular geometry, and visual noise—human performance remains perfect (100\% across all categories). Zero-shot VLM performance, however, degrades substantially. Success rates fall to 80.95\% in single-plant multi-target tasks (Row 2 in Table \ref{tab:complex.polyculture})and to 27.27\% and 31.58\% in multi-plant conditions (Rows 3 and 4 Table \ref{tab:complex.polyculture}). SPL similarly decreases, and tool call counts increase significantly, indicating repeated attempts to re-establish grounding and recover from planning failures.

These results illustrate that complex visual structure and longer horizon tasks impose a strong burden on the VLM’s spatial reasoning and perception-action grounding. Unlike human operators, who leverage contextual understanding and adaptive heuristics, the VLM struggles to maintain consistent target tracking and to execute coherent long-horizon sequences in cluttered environments. Importantly, these failures are unlikely to stem solely from context window limitations. Rather, they also reflect the absence of explicit task memory and structured planning mechanisms. In our current system, the VLM operates primarily as a reactive high-level decision module, selecting actions based on current observations, instructions, and limited interaction history, without maintaining an explicit representation of task progress.

This limitation suggests two promising directions for improving long-horizon performance. First, augmenting the system with structured memory—beyond the existing semantic map—such as explicit task-state representations could enable the agent to track visited regions, inspected plants, and completed subtasks, improving consistency and reducing redundant exploration. Second, incorporating symbolic or hierarchical planning could provide stronger guarantees on task decomposition, ordering, and safety, enabling more reliable execution in complex agricultural environments. Together, these additions point toward hybrid architectures that combine VLM perception with explicit memory and planning as a path forward for robust long-horizon autonomy.

\subsection{Results Under Noisy and Variant Conditions}

Tables \ref{tab:ablation.polar.actions} and \ref{tab:ablation.robotcentric} present the results obtained with noisy semantic maps. Compared to Table \ref{tab:main.results}, which uses ground-truth maps, the zero-shot VLM agent experiences a drop of approximately 12\% in success rate and 26\% in completion rate. This can be attributed to an increase in the number of failures due to map noise. Specifically, looking at the failure statistics presented in Fig. \ref{fig:failure_statistics}, the total number of failures under noisy conditions are approximately twice the number of failures in ideal conditions. The main reason for this performance drop are target misclassifications and navigation failures (the latter categorized under miscellaneous behaviors) driven by false positives in the map. When the robot approaches a false-positive plant, the VLM is often overconfident and incorrectly assumes that the observed plant corresponds to the target, leading to object misclassification. Interestingly, this can occur even between visually distinct plants such as eggplants and bell peppers. 
Additionally, when the VLM detects a mismatch between the observed plant and the target, it attempts alternative exploration strategies (e.g., turning in place or searching for another location in the map). However, in most cases, it still fails to locate the correct target, resulting in navigation failures. In some cases, the VLM detects the true target in the background and proceeds to capture “close-up” views despite being visually implausible.

The results in Tables \ref{tab:ablation.polar.actions} and \ref{tab:ablation.robotcentric} do not indicate a clear advantage for the VLM-few-shot agent over its zero-shot counterpart. In some task categories, the few-shot agent even exhibits lower success and completion rates. On average across categories, the zero-shot VLM achieves a 7\% higher success rate than the few-shot agent.  By examining the demonstrations used in the few-shot setting, we found that although they include examples addressing the main failure modes produced by noisy maps, their diversity may be insufficient to promote robust behavior, particularly in the presence of false positives. Additionally, since these demonstrations included textual descriptions only for efficient token usage, they might lack proper grounding.  Increasing the diversity of demonstrations to better identify and handle false positives, as well as incorporating images into the demonstrations may improve performance. Additionally, explicitly providing object confidence scores to the VLM, along with updating the map based on VLM observations, could improve robustness under noisy conditions.

Furthermore, across multiple task categories, the agent using the front camera with overlaid polar actions (Table \ref{tab:ablation.polar.actions}) achieves, on average, a 7\% higher success and completion rate than the agent using a robot-centric map (Table \ref{tab:ablation.robotcentric}). This may be because VLMs are not necessarily trained with top-down semantic occupancy maps, making image-space planning more intuitive for them. Fine-tuning VLMs with 2D semantic occupancy maps may be necessary to fully leverage their spatial and semantic information.

\begin{table*}[htp!]
\centering
\caption{\textbf{Noisy Map Study Results (Complex Polyculture Environment, Noisy Map, Polar Actions)}}
\resizebox{\textwidth}{!}{%
    \begin{tabular}{l@{\hskip 4mm} c c c c c c c c c}
    \toprule
Task Category & Operator & Number of Trials & Success Rate (\%) $\uparrow$ & Number of Tool Calls $\downarrow$ & Task Completion Rate (\%) $\uparrow$ & SPL $\uparrow$\\
    \midrule
    \multirow{3}{*}{Single Plant - Single Target}
        & Human & 7 & 100.00 & 12.43$\pm$4.72 & 100.00$\pm$0.00 & 100.00\\
        & Zero-Shot VLM & 7 & 85.71 & 10.17$\pm$1.46 & 85.71$\pm$34.99 & 83.90\\
        & Few-Shot VLM & 7 & 85.71 & 10.17$\pm$1.34 & 85.71$\pm$34.99 & 79.30\\
    \midrule
    \multirow{3}{*}{Single Plant - Multiple Target}
        & Human & 7 & 100.00 & 15.86$\pm$2.75 & 100.00$\pm$0.00 & 100.00\\
        & Zero-Shot VLM & 7 & 28.57 & 21.50$\pm$3.50 & 53.57$\pm$38.80 & 28.57\\
        & Few-Shot VLM & 7 & 42.86 & 19.00$\pm$2.45 & 75.00$\pm$32.73 & 42.37\\
    \midrule
    \multirow{3}{*}{Multiple Plant - Single Target}
        & Human & 7 & 100.00 & 28.14$\pm$1.73 & 100.00$\pm$0.00 & 100.00\\
        & Zero-Shot VLM & 7 & 42.86 & 31.00$\pm$5.66 & 81.00$\pm$18.67 & 32.91\\
        & Few-Shot VLM & 7 & 14.29 & 31.00$\pm$0.00 & 52.43$\pm$30.25 & 13.49\\
    \midrule
    \multirow{3}{*}{Multiple Plant - Multiple Target}
        & Human & 7 & 100.00 & 35.43$\pm$5.39 & 100.00$\pm$0.00 & 100.00\\
        & Zero-Shot VLM & 7 & 28.57 & 37.50$\pm$8.50 & 52.00$\pm$34.12 & 18.74\\
        & Few-Shot VLM & 7 & 0.00 & NA & 45.86$\pm$29.78 & 0.00\\
    \midrule
    \multirow{3}{*}{Average Over All Categories}
        & Human & 28 & 100.00 & 22.96$\pm$3.65 & 100.00$\pm$0.00 & 100.00\\
        & Zero-Shot VLM & 28 & 46.43 & 25.04$\pm$4.78 & 68.07$\pm$31.64 & 41.03\\
        & Few-Shot VLM & 28 & 35.72 & 20.06$\pm$1.26 & 64.75$\pm$31.94 & 33.79\\
    \bottomrule
    \end{tabular}}
\label{tab:ablation.polar.actions}
\end{table*}

\begin{table*}[htp!]
\centering
\caption{\textbf{Noisy Map Study Results (Complex Polyculture Environment, Noisy Map, Robot-Centric Map)}}
\resizebox{\textwidth}{!}{%
    \begin{tabular}{l@{\hskip 4mm} c c c c c c c c c}
    \toprule
Task Category & Operator & Number of Trials & Success Rate (\%) $\uparrow$ & Number of Tool Calls $\downarrow$ & Task Completion Rate (\%) $\uparrow$ & SPL $\uparrow$\\
    \midrule
    \multirow{3}{*}{Single Plant - Single Target}
        & Human & 7 & 100.00 & 11.29$\pm$2.25 & 100.00$\pm$0.00 & 100.00\\
        & Zero-Shot VLM & 7 & 57.14 & 11.75$\pm$2.86 & 64.29$\pm$44.03 & 44.07\\
        & Few-Shot VLM & 7 & 71.43 & 9.80$\pm$0.98 & 78.57$\pm$36.42 & 55.71\\
    \midrule
    \multirow{3}{*}{Single Plant - Multiple Target}
        & Human & 7 & 100.00 & 16.43$\pm$1.40 & 100.00$\pm$0.00 & 100.00\\
        & Zero-Shot VLM & 7 & 42.86 & 19.00$\pm$2.16 & 50.00$\pm$46.29 & 29.88\\
        & Few-Shot VLM & 7 & 57.14 & 16.75$\pm$0.83 & 67.86$\pm$43.74 & 48.68\\
    \midrule
    \multirow{3}{*}{Multiple Plant - Single Target}
        & Human & 7 & 100.00 & 30.29$\pm$4.16 & 100.00$\pm$0.00 & 100.00\\
        & Zero-Shot VLM & 7 & 14.29 & 38.00$\pm$0.00 & 49.86$\pm$28.14 & 11.28\\
        & Few-Shot VLM & 7 & 0.00 & NA & 57.29$\pm$25.00 & 0.00\\
    \midrule
    \multirow{3}{*}{Multiple Plant - Multiple Target}
        & Human & 7 & 100.00 & 34.29$\pm$6.09 & 100.00$\pm$0.00 & 100.00\\
        & Zero-Shot VLM & 7 & 42.86 & 44.00$\pm$14.45 & 76.29$\pm$23.36 & 35.87\\
        & Few-Shot VLM & 7 & 0.00 & NA & 49.57$\pm$19.27 & 0.00\\
    \midrule
    \multirow{3}{*}{Average Over All Categories}
        & Human & 28 & 100.00 & 23.08$\pm$3.48 & 100.00$\pm$0.00 & 100.00\\
        & Zero-Shot VLM & 28 & 39.29 & 28.19$\pm$4.87 & 60.11$\pm$35.46 & 30.28\\
        & Few-Shot VLM & 28 & 32.14 & 13.28$\pm$0.90 & 63.32$\pm$31.11 & 26.10\\
    \bottomrule
    \end{tabular}}
\label{tab:ablation.robotcentric}
\end{table*}

\subsection{Real-World Experiments}

\begin{table*}[htp!]
\centering
\caption{\textbf{Main Results Real-World Experiments}}
\resizebox{\textwidth}{!}{%
    \begin{tabular}{l@{\hskip 4mm} c c c c c c c c c}
    \toprule
Task Category & Operator & Number of Trials & Success Rate (\%) $\uparrow$ & Number of Tool Calls $\downarrow$ & Task Completion Rate (\%) $\uparrow$ & SPL $\uparrow$\\
    \midrule
    \multirow{2}{*}{Single Plant - Single Target} & Human & 10 & 100.0 & 7.60$\pm$0.66 & 100.0$\pm$0.0 & 100.0\\
    & Zero-Shot VLM & 10 & 90.0 & 10.33$\pm$1.63 & 95.0$\pm$15.00 & 65.16\\
    \midrule
    \multirow{2}{*}{Single Plant - Multiple Target} & Human & 9 & 100.0 & 13.22$\pm$1.40 & 100.0$\pm$0.0 & 100.0\\
    & Zero-Shot VLM & 9 & 77.78 & 16.86$\pm$2.29 & 93.94$\pm$12.45 & 63.78\\
    \midrule
    \multirow{2}{*}{Multiple Plant - Single Target} & Human & 8 & 100.0 & 24.38$\pm$0.86 & 100.0$\pm$0.0 & 100.0\\
    & Zero-Shot VLM  & 10 & 40.00 & 32.25$\pm$2.38 & 88.20$\pm$10.68 & 33.02\\
    \midrule
    \multirow{2}{*}{Multiple Plant - Multiple Target} & Human & 7 & 100.0 & 34.71$\pm$4.92 & 100.0$\pm$0.0 & 100.0\\
    & Zero-Shot VLM & 9 & 0.00 & NA & 78.78$\pm$14.68 & 0.0\\

    \bottomrule
    \end{tabular}}
\label{tab:real-world-results}
\end{table*}

\begin{figure*}[htp]
    \centering
    \includegraphics[width=1\textwidth]{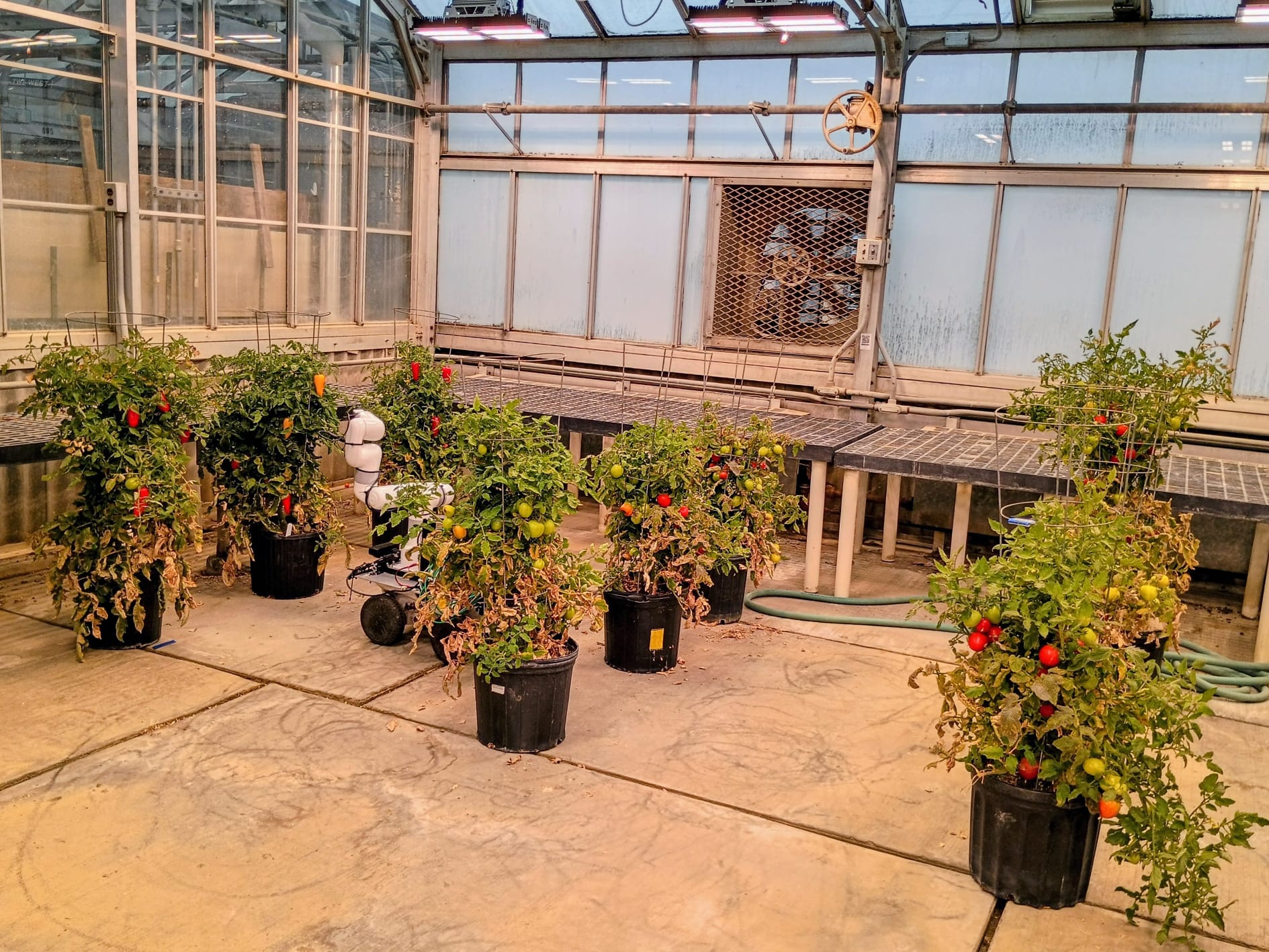}
    \caption{\textbf{Experimental setup.} Real-world testing of our method was done using a mobile manipulator in a greenhouse environment with rows of red peppers and tomatoes in potted plants.}

    \label{fig:experimental_setup}
\end{figure*}

We evaluate our framework in a real greenhouse environment. We used a mobile manipulator consisting of a 6-DoF robotic arm mounted on a Terrasentia ground robot, closely resembling our simulation platform. The robot is equipped with two stereo cameras: one mounted on the robot base (front-facing) and another on the manipulator’s end effector. For localization, we run stereo-inertial SLAM. The full framework is deployed on a Jetson Orin platform mounted on the robot. Our greenhouse setup consists of three crop rows, including bell pepper and tomato plants grown in pots, as shown in Fig. \ref{fig:experimental_setup}
We begin by constructing a semantic occupancy map through manual teleoperation of the robot. We conducted approximately ten trials for each task category using both the zero-shot VLM and a human operator. To ensure reproducibility and enable a fair comparison, we kept the occupancy information fixed and replaced the noisy object positions with their ground-truth values. A demonstration video and the metadata of all the episodes are provided in the project website.

The quantitative results are presented in Table \ref{tab:real-world-results}. Interestingly, the performance of our approach in the real-world environment closely matches that observed in simulation across all evaluation metrics (Table \ref{tab:main.results}). We attribute this result to two main factors. First, the proposed semantic map representation is largely domain-agnostic and can be transferred seamlessly between simulated and real environments. Second, modern VLMs exhibit strong generalization capabilities across both synthetic and real camera imagery. As a result, the agent demonstrates remarkably similar behaviors in simulation and in the real world.


Beyond confirming sim-to-real transfer, our real-world experiments surface several important insights.

Robust localization remains one of the main challenges. Unlike the simulation environment, which provides perfect ground-truth robot poses, real-world localization relies on visual-inertial SLAM and is subject to drift over time, particularly when loop closures are not detected. This often leads to navigation failures.

Regarding active vision performance, the VLM is capable of capturing close-up inspections of different targets. However, it exhibits behaviors similar to those observed in simulation, such as target misclassification, repetitive actions, and incorrect self-evaluation. These issues result in failures, including capturing background objects or getting stuck in infinite loops.

Overall, our experimental results suggest that the proposed framework is promising for flexible and user-friendly crop monitoring tasks. However, further research is needed to improve robustness to perception noise and localization errors, as well as to enhance efficiency in selecting optimal inspection targets, as well as perception and action tools.

\subsection{Failure Cases}
\label{sec:failure_cases}

\begin{figure*}[htp]
    \centering
    \includegraphics[width=1\textwidth]{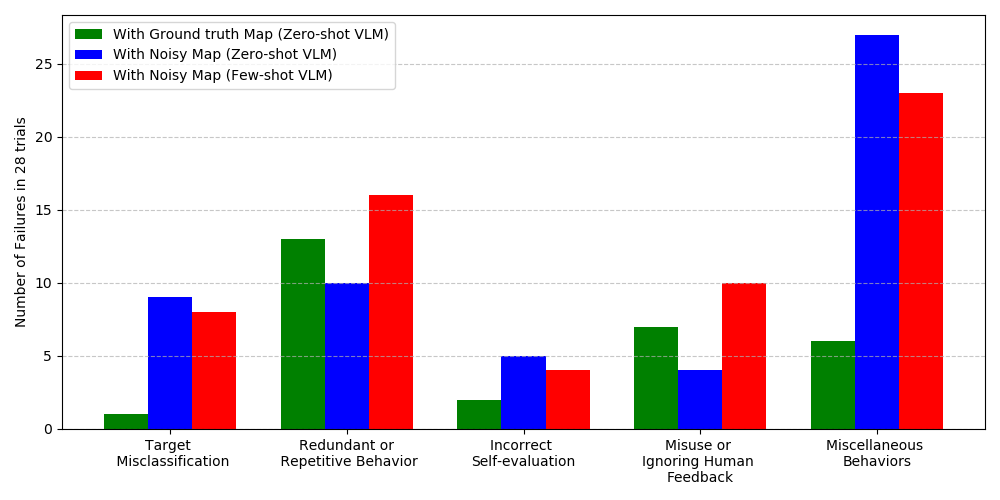}
    \caption{\textbf{Failure statistics computed from 28 trials for each condition: zero-shot VLM with a ground-truth map, zero-shot VLM with a noisy map, and few-shot VLM with a noisy map.} Under ideal conditions (ground-truth map), redundant or repetitive behaviors are the dominant failure mode. In contrast, when using noisy maps, target misclassification and miscellaneous behaviors become more prevalent. Additionally, the few-shot setting does not lead to a clear reduction in the overall number of failures.
Total number of failures for each setting: 29 with zero-shot VLM with ground truth map, 55 with zero-shot VLM with noisy map, and 61 with few-shot VLM with noisy map.
}

    \label{fig:failure_statistics}
\end{figure*}

Fig. \ref{fig:failure_statistics} shows the distribution of failure types across 28 trials (7 per task category) for each agent mode, including zero-shot and few-shot VLM under both ideal and noisy conditions.

\textbf{Failure Mode 1: Target Misclassification}

The VLM frequently demonstrated errors in basic object categorization, leading to downstream failures in navigation and active vision. These errors included confusing visually similar produce—such as interpreting yellow tomatoes as ripe or green tomatoes—and misidentifying red bell peppers or tomatoes as strawberries. In some cases, the model incorrectly distinguished between plant species altogether. These behaviors become more frequent under noisy conditions, as the VLM remains overconfident in the map despite the presence of false-positive objects. These failures indicate that, despite access to high-resolution imagery and semantic map metadata, the model may rely on unstable visual heuristics or incomplete contextual reasoning. Such misclassifications often propagated through the control loop, producing incorrect subgoal selection or premature task termination.

Beyond broad object class errors, the VLM struggled to correctly interpret fine-grained attributes such as color, plant part, or ripeness level. Sometimes, red tomatoes were labeled "less ripe" or selected the wrong instance within a cluster when multiple visually similar options were present. Attribute-specific errors were especially common for tasks requiring differentiation among leaves, stems, and fruit, where the model occasionally centered on or captured images of the wrong part despite the correct object being in view. These issues suggest that attribute-based recognition remains a weak point, even when class-level recognition appears correct.

\textbf{Failure Mode 2: Redundant or Repetitive Behavior}
A recurring behavioral pattern involved the model repeatedly centering the same target without making forward progress toward approaching or capturing it. In several instances, the VLM entered loops where it continually issued centering commands, apparently satisfied with the visual framing yet failing to advance to the next step of the task. This also occurred during image capture, where the model would repeatedly reframe a target it had already documented, or unnecessarily reacquire the same viewpoint from nearly identical camera poses. Such patterns indicate a disconnect between visual satisfaction criteria and the task-level state machine designed to govern completion.

The model also exhibited repetitive navigation behavior, reselecting the same plant or the same goal location after completion. In some episodes, it captured the same image twice due to relying solely on the most recent camera frame, effectively forgetting previously accomplished subtasks. In others, it navigated back to a plant it had already visited, or misinterpreted a previously achieved goal as the next required target. These loops reflect inconsistencies in temporal reasoning and insufficient handling of task memory within the prompt or system context.

\textbf{Failure Mode 3: Incorrect Self-Evaluation}
 Several failures stemmed from the model’s inability to correctly evaluate the results of its own actions. In several episodes, the VLM believed it had successfully completed a subgoal even when the captured image did not contain the required element. Conversely, it sometimes assumed it had failed despite having captured the correct target, prompting further unnecessary exploration or repeated re-execution of a completed step. These errors occasionally triggered cascades of redundant behavior, such as revisiting plants or recentering previously captured objects. Even when user feedback was provided, the model would sometimes integrate it incorrectly or continue to question its progress. These findings highlight fundamental challenges in perception-action grounding, particularly in interpreting visual feedback in the context of multi-stage tasks.

\textbf{Failure Mode 4: Misuse or Ignoring Human Feedback}
Although the system is designed to leverage human input for correction and confirmation, the VLM often misused or misinterpreted this feedback channel. At times, the model ignored user prompts to proceed to the next subgoal, continuing to refine an already complete image. In other cases, it requested confirmation at inappropriate times—such as immediately after completing a subgoal or before attempting the next action—leading to unnecessary slowdowns. The model occasionally asked for guidance despite having sufficient information to continue independently, or applied user feedback in an overly literal or incorrect manner. These behaviors illustrate a lack of reliable grounding for human-provided signals, and highlight broader challenges in building VLM-driven systems that must balance autonomy with interactive correction.

\textbf{Failure Mode 5: Miscellaneous Behaviors}
A variety of additional behaviors fell outside the primary failure categories but remain informative for understanding the system’s overall robustness. These include over-collecting images due to ambiguous task interpretation, completing goals out of order, or extracting an incorrect number of subgoals from natural-language instructions. The model sometimes relied on stale or noisy global map information instead of using available local camera feedback, or hallucinated plant locations entirely. Navigation failures were also common, such as getting stuck, approaching the wrong plant, moving backward unnecessarily, or attempting to explore nonexistent vegetation. The VLM occasionally neglected to use the robot-centric map for local navigation or misused polar actions when available. Navigation failures become more prominent under noisy conditions. When the VLM detects a mismatch between the observed plant and the target, it attempts alternative exploration strategies (e.g., turning in place or searching for another location in the map). However, in most cases, it still fails to locate the correct target due to poor local navigation and over-reliance on a noisy map. Despite these failures, some episodes showcased unexpectedly competent behaviors, such as clever backward camera use or successful fallback to the local map after a global navigation failure. These mixed outcomes underscore both the promise and instability of current VLM control in structured agricultural environments.

\section{Limitations}
\textbf{The VLM agent performs well on short-horizon tasks under noiseless conditions}. While the VLM agent achieves performance close to that of humans on short-horizon tasks, its performance deteriorates on long-horizon tasks. This limitation is also evident when providing demonstrations to the VLM agent: the model often fails to capture fine-grained details or desired behaviors provided in the demonstrations. One possible explanation for this performance drop is the longer context window required by long-horizon tasks, which remains an active research challenge for both LLMs and VLMs. In addition, in our implementation, the VLM primarily operates as a high-level decision module that selects actions based on current observations, instructions, and conversation history, without maintaining an explicit representation of task progress. As a result, performance degradation may also stem from the lack of explicit task memory and structured planning. Integrating structured memory beyond the semantic map (e.g., task-state representations) could enable the agent to track inspected plants or visited regions, improving consistency over extended missions. Similarly, incorporating symbolic or hierarchical planning could enhance safety and provide stronger guarantees for task decomposition and execution order in long-horizon crop monitoring tasks.

The performance degradation is exacerbated when using noisy maps. In many cases, the VLM does not even detect false-positive objects in the map. These misdetections may be attributed to the overconfidence commonly exhibited by VLMs, as reported in prior work \cite{groot2024overconfidence}. However, they may also stem from dataset bias, since the model is evaluated in a simulated environment whose visual appearance differs significantly from the real images used during training.

\textbf{The VLM agent selects suboptimal spatial goals}. Crop-monitoring tasks require the ability to reason about proximity to objects, object size, occlusions, and the spatial relationships between objects and the robot. While humans excel at these skills, the VLM agent shows substantial deficiencies, leading to failures and inefficient task execution.  For example, during navigation the agent occasionally selects intermediate goals that are feasible but suboptimal, resulting in increased traversal time or inspecting a previously visited plant. Similarly, during perception-driven alignment, the agent may require multiple attempts to correctly select and center the intended target object in the image. These behaviors indicate limitations in spatial reasoning consistency and short-term task memory, which can compound over extended task sequences and ultimately degrade overall task performance. 

A key reason for this limitation is that most VLMs are trained primarily on image–text pairs, where spatial relationships are rarely annotated explicitly. As a result, the model learns strong associations between objects and their semantic labels but develops only weak representations of geometric structure, distance, and 3D layout. Furthermore, because VLMs operate on static 2D images, they lack an explicit notion of depth or egocentric spatial context. Fine-tuning the model on spatial-reasoning tasks may help improve the VLM agent’s performance in crop monitoring. Incorporating additional training signals that capture embodied spatial interaction, such as video sequences or geometric scene representations, may help improve the consistency of spatial decision-making in similar crop-monitoring tasks.

\textbf{The reliance on large-scale foundation models introduces significant latency, limiting the system's applicability in time-critical robotic operations}. We found that the average time between decision steps across all trials was 5.4 seconds, of which 1.8 seconds are attributed to VLM inference. The high inference cost necessitates the use of sparse image observations rather than continuous visual streams, which forces the agent to make decisions based on snapshot data that may miss dynamic environmental cues. A lightweight VLM would not only reduce inference time, enabling smoother real-time control, but could also support high-frequency video input. Processing video rather than static frames would provide the model with richer temporal and 3D cues, potentially resolving the spatial ambiguity and context grounding issues observed in our experiments.

\textbf{Simulation to Reality Transfer}

Overall, we find that the high-level structure of our framework transfers well to the real world. The VLM is able to interpret visual inputs, decompose tasks, and invoke action primitives in a zero-shot setting, demonstrating that the core paradigm of language-conditioned, vision-driven task execution is viable beyond simulation. In particular, the system successfully performs short-horizon inspection tasks and leverages active vision to gather task-relevant observations.

However, our experiments highlight several key challenges that impact end-to-end performance. Most notably, robust localization emerges as a primary bottleneck. Unlike simulation, which provides perfect state information, real-world deployment relies on stereo-inertial SLAM, which is susceptible to drift—especially in the absence of reliable loop closures. This drift frequently leads to navigation errors and task failures, indicating that system performance is tightly coupled with the reliability of the underlying localization stack.

In addition, perception and reasoning errors in the VLM persist and are amplified in the real world. Similar to simulation, we observe target misclassification, incorrect subtask verification, and repetitive action selection. In practice, these issues manifest as failure modes such as inspecting irrelevant background regions or entering execution loops without task completion. These results suggest that while the VLM is capable of grounding and reasoning over real visual inputs, its failure modes remain a limiting factor for robust autonomy.

More broadly, real-world perception introduces compounding noise that is absent in simulation. Sensor noise, imperfect detections, and partial observability affect both the inputs to the VLM and the downstream execution of actions. While plant variability and environmental complexity do introduce distribution shifts, our results indicate that these are not the primary limiting factors. Instead, the interaction between perception noise, VLM reasoning errors, and localization drift plays a more significant role in system failures.

These findings suggest several important directions for future work. Improving localization robustness, particularly in repetitive agricultural environments, is critical for reliable deployment. Additionally, enhancing the VLM’s ability to recover from errors—through better self-evaluation, loop prevention, or uncertainty-aware reasoning—will be essential for stable long-horizon execution. Finally, tighter integration between perception, mapping, and decision-making may help mitigate error propagation across the system.




\section{Conclusion and Future Work}

In this work, we introduced a novel, modular framework for visual-language-guided task planning in horticultural environments. By interleaving input-source queries with a library of action primitives, our system enables mobile manipulators to perform flexible crop monitoring tasks using natural language instructions. We integrated open-vocabulary semantic occupancy maps and enriched camera observations to ground high-level reasoning in the physical environment. 
Our extensive evaluation in simulated monoculture and polyculture environments reveals a distinct dichotomy in current VLM capabilities for agricultural robotics. We found that the zero-shot VLM performs robustly on short-horizon, single-target tasks, achieving success rates comparable to human operators. However, performance degrades significantly as task complexity increases. In long-horizon scenarios involving multiple plants and targets, the VLM struggled with spatial reasoning, efficient navigation, and context management, often resulting in redundant behaviors and lower success rates compared to human expert trials. Furthermore, our study demonstrated that system reliability is heavily dependent on map accuracy, with performance dropping notably when facing perception noise.

To bridge the gap between current capabilities and field-ready deployment, our future work will focus on addressing these spatial and temporal limitations. We aim to fine-tune small VLMs on agricultural datasets enriched with explicit spatial relationships and map data to enhance geometric reasoning. Additionally, we plan to incorporate richer input modalities, such as video streams or 3D scene representations, to improve depth perception and temporal understanding. Finally, to mitigate the performance degradation observed in long-horizon tasks, we will explore advanced memory management techniques to overcome context window constraints and prevent the repetitive behaviors observed in our experiments.

\bibliographystyle{IEEEtran}
\bibliography{mybibliography}

@inproceedings{lehnert20193d,
  title={3d move to see: Multi-perspective visual servoing towards the next best view within unstructured and occluded environments},
  author={Lehnert, Chris and Tsai, Dorian and Eriksson, Anders and McCool, Chris},
  booktitle={2019 IEEE/RSJ International Conference on Intelligent Robots and Systems (IROS)},
  pages={3890--3897},
  year={2019},
  organization={IEEE}
}

@inproceedings{zaenker2021viewpoint,
  title={Viewpoint planning for fruit size and position estimation},
  author={Zaenker, Tobias and Smitt, Claus and McCool, Chris and Bennewitz, Maren},
  booktitle={2021 IEEE/RSJ International Conference on Intelligent Robots and Systems (IROS)},
  pages={3271--3277},
  year={2021},
  organization={IEEE}
}

@inproceedings{pan2023panoptic,
  title={Panoptic mapping with fruit completion and pose estimation for horticultural robots},
  author={Pan, Yue and Magistri, Federico and L{\"a}be, Thomas and Marks, Elias and Smitt, Claus and McCool, Chris and Behley, Jens and Stachniss, Cyrill},
  booktitle={2023 IEEE/RSJ International Conference on Intelligent Robots and Systems (IROS)},
  pages={4226--4233},
  year={2023},
  organization={IEEE}
}

@article{hornung2013octomap,
  title={OctoMap: An efficient probabilistic 3D mapping framework based on octrees},
  author={Hornung, Armin and Wurm, Kai M and Bennewitz, Maren and Stachniss, Cyrill and Burgard, Wolfram},
  journal={Autonomous robots},
  volume={34},
  pages={189--206},
  year={2013},
  publisher={Springer}
}

@inproceedings{werby2024hierarchical,
  title={Hierarchical open-vocabulary 3d scene graphs for language-grounded robot navigation},
  author={Werby, Abdelrhman and Huang, Chenguang and B{\"u}chner, Martin and Valada, Abhinav and Burgard, Wolfram},
  booktitle={First Workshop on Vision-Language Models for Navigation and Manipulation at ICRA 2024},
  year={2024}
}

@article{liang2022code,
  title={Code as policies: Language model programs for embodied control},
  author={Liang, Jacky and Huang, Wenlong and Xia, Fei and Xu, Peng and Hausman, Karol and Ichter, Brian and Florence, Pete and Zeng, Andy},
  journal={arXiv preprint arXiv:2209.07753},
  year={2022}
}

@article{huang2022inner,
  title={Inner monologue: Embodied reasoning through planning with language models},
  author={Huang, Wenlong and Xia, Fei and Xiao, Ted and Chan, Harris and Liang, Jacky and Florence, Pete and Zeng, Andy and Tompson, Jonathan and Mordatch, Igor and Chebotar, Yevgen and others},
  journal={arXiv preprint arXiv:2207.05608},
  year={2022}
}

@article{huang2022visual,
  title={Visual language maps for robot navigation},
  author={Huang, Chenguang and Mees, Oier and Zeng, Andy and Burgard, Wolfram},
  journal={arXiv preprint arXiv:2210.05714},
  year={2022}
}

@inproceedings{shah2023lm,
  title={Lm-nav: Robotic navigation with large pre-trained models of language, vision, and action},
  author={Shah, Dhruv and Osi{\'n}ski, B{\l}a{\.z}ej and Levine, Sergey and others},
  booktitle={Conference on robot learning},
  pages={492--504},
  year={2023},
  organization={PMLR}
}

@article{jiang2024roboexp,
  title={Roboexp: Action-conditioned scene graph via interactive exploration for robotic manipulation},
  author={Jiang, Hanxiao and Huang, Binghao and Wu, Ruihai and Li, Zhuoran and Garg, Shubham and Nayyeri, Hooshang and Wang, Shenlong and Li, Yunzhu},
  journal={arXiv preprint arXiv:2402.15487},
  year={2024}
}

@article{chen2024mapgpt,
  title={Mapgpt: Map-guided prompting with adaptive path planning for vision-and-language navigation},
  author={Chen, Jiaqi and Lin, Bingqian and Xu, Ran and Chai, Zhenhua and Liang, Xiaodan and Wong, Kwan-Yee K},
  journal={arXiv preprint arXiv:2401.07314},
  year={2024}
}

@article{zhong2024topv,
  title={Topv-nav: Unlocking the top-view spatial reasoning potential of mllm for zero-shot object navigation},
  author={Zhong, Linqing and Gao, Chen and Ding, Zihan and Liao, Yue and Ma, Huimin and Zhang, Shifeng and Zhou, Xu and Liu, Si},
  journal={arXiv preprint arXiv:2411.16425},
  year={2024}
}

@article{wu2024voronav,
  title={Voronav: Voronoi-based zero-shot object navigation with large language model},
  author={Wu, Pengying and Mu, Yao and Wu, Bingxian and Hou, Yi and Ma, Ji and Zhang, Shanghang and Liu, Chang},
  journal={arXiv preprint arXiv:2401.02695},
  year={2024}
}

@article{nasiriany2024pivot,
  title={Pivot: Iterative visual prompting elicits actionable knowledge for vlms},
  author={Nasiriany, Soroush and Xia, Fei and Yu, Wenhao and Xiao, Ted and Liang, Jacky and Dasgupta, Ishita and Xie, Annie and Driess, Danny and Wahid, Ayzaan and Xu, Zhuo and others},
  journal={arXiv preprint arXiv:2402.07872},
  year={2024}
}

@article{elnoor2024robot,
  title={Robot navigation using physically grounded vision-language models in outdoor environments},
  author={Elnoor, Mohamed and Weerakoon, Kasun and Seneviratne, Gershom and Xian, Ruiqi and Guan, Tianrui and Jaffar, Mohamed Khalid M and Rajagopal, Vignesh and Manocha, Dinesh},
  journal={arXiv preprint arXiv:2409.20445},
  year={2024}
}

@article{haghighat2025multimodal,
  title={Multimodal Language Models in Agriculture: A Tutorial and Survey},
  author={Haghighat, Mohammadreza and Saleh, Alzayat and Azghadi, Mostafa Rahimi},
  journal={Authorea Preprints},
  year={2025},
  publisher={Authorea}
}

@article{zuzuarregui2025leveraging,
  title={Leveraging LLMs for mission planning in precision agriculture},
  author={Zuzu{\'a}rregui, Marcos Abel and Carpin, Stefano},
  journal={arXiv preprint arXiv:2506.10093},
  year={2025}
}

@article{ahn2022can,
  title={Do as i can, not as i say: Grounding language in robotic affordances},
  author={Ahn, Michael and Brohan, Anthony and Brown, Noah and Chebotar, Yevgen and Cortes, Omar and David, Byron and Finn, Chelsea and Fu, Chuyuan and Gopalakrishnan, Keerthana and Hausman, Karol and others},
  journal={arXiv preprint arXiv:2204.01691},
  year={2022}
}

@inproceedings{huang2022language,
  title={Language models as zero-shot planners: Extracting actionable knowledge for embodied agents},
  author={Huang, Wenlong and Abbeel, Pieter and Pathak, Deepak and Mordatch, Igor},
  booktitle={International conference on machine learning},
  pages={9118--9147},
  year={2022},
  organization={PMLR}
}

@inproceedings{kaelbling2011hierarchical,
  title={Hierarchical task and motion planning in the now},
  author={Kaelbling, Leslie Pack and Lozano-P{\'e}rez, Tom{\'a}s},
  booktitle={2011 IEEE international conference on robotics and automation},
  pages={1470--1477},
  year={2011},
  organization={IEEE}
}

@inproceedings{wolfe2010combined,
  title={Combined task and motion planning for mobile manipulation},
  author={Wolfe, Jason and Marthi, Bhaskara and Russell, Stuart},
  booktitle={Proceedings of the International Conference on Automated Planning and Scheduling},
  volume={20},
  pages={254--257},
  year={2010}
}

@article{fikes1971strips,
  title={STRIPS: A new approach to the application of theorem proving to problem solving},
  author={Fikes, Richard E and Nilsson, Nils J},
  journal={Artificial intelligence},
  volume={2},
  number={3-4},
  pages={189--208},
  year={1971},
  publisher={Elsevier}
}

@inproceedings{zhang2022task,
  title={Task and motion planning methods: applications and limitations},
  author={Zhang, Kai and Lucet, Eric and Sandretto, Julien Alexandre Dit and Kchir, Selma and Filliat, David},
  booktitle={ICINCO 2022-19th International Conference on Informatics in Control, Automation and Robotics},
  pages={476--483},
  year={2022},
  organization={SCITEPRESS-Science and Technology Publications}
}

@article{getahun2024application,
  title={Application of precision agriculture technologies for sustainable crop production and environmental sustainability: A systematic review},
  author={Getahun, Sewnet and Kefale, Habtamu and Gelaye, Yohannes},
  journal={The Scientific World Journal},
  volume={2024},
  number={1},
  pages={2126734},
  year={2024},
  publisher={Wiley Online Library}
}

@article{spagnuolo2025agricultural,
  title={Agricultural Robotics: A Technical Review Addressing Challenges in Sustainable Crop Production},
  author={Spagnuolo, Maria and Todde, Giuseppe and Caria, Maria and Furnitto, Nicola and Schillaci, Giampaolo and Failla, Sabina},
  journal={Robotics},
  volume={14},
  number={2},
  pages={9},
  year={2025},
  publisher={MDPI}
}

@inproceedings{cuaran2025active,
  title={Active Semantic Mapping with Mobile Manipulator in Horticultural Environments},
  author={Cuaran, Jose and Ahluwalia, Kulbir Singh and Koe, Kendall and Uppalapati, Naveen Kumar and Chowdhary, Girish},
  booktitle={2025 IEEE International Conference on Robotics and Automation (ICRA)},
  pages={12716--12722},
  year={2025},
  organization={IEEE}
}

@inproceedings{freeman2024autonomous,
  title={Autonomous apple fruitlet sizing with next best view planning},
  author={Freeman, Harry and Kantor, George},
  booktitle={2024 IEEE International Conference on Robotics and Automation (ICRA)},
  pages={15847--15853},
  year={2024},
  organization={IEEE}
}

@article{hu2023look,
  title={Look before you leap: Unveiling the power of gpt-4v in robotic vision-language planning},
  author={Hu, Yingdong and Lin, Fanqi and Zhang, Tong and Yi, Li and Gao, Yang},
  journal={arXiv preprint arXiv:2311.17842},
  year={2023}
}

@article{ginting2025enter,
  title={Enter the Mind Palace: Reasoning and Planning for Long-term Active Embodied Question Answering},
  author={Ginting, Muhammad Fadhil and Kim, Dong-Ki and Meng, Xiangyun and Reinke, Andrzej and Krishna, Bandi Jai and Kayhani, Navid and Peltzer, Oriana and Fan, David D and Shaban, Amirreza and Kim, Sung-Kyun and others},
  journal={arXiv preprint arXiv:2507.12846},
  year={2025}
}

@inproceedings{majumdar2024openeqa,
  title={Openeqa: Embodied question answering in the era of foundation models},
  author={Majumdar, Arjun and Ajay, Anurag and Zhang, Xiaohan and Putta, Pranav and Yenamandra, Sriram and Henaff, Mikael and Silwal, Sneha and Mcvay, Paul and Maksymets, Oleksandr and Arnaud, Sergio and others},
  booktitle={Proceedings of the IEEE/CVF conference on computer vision and pattern recognition},
  pages={16488--16498},
  year={2024}
}

@article{lan2025experience,
  title={Experience is the Best Teacher: Grounding VLMs for Robotics through Self-Generated Memory},
  author={Lan, Guowei and Qu, Kaixian and Zurbr{\"u}gg, Ren{\'e} and Chen, Changan and Mower, Christopher E and Bou-Ammar, Haitham and Hutter, Marco},
  journal={arXiv preprint arXiv:2507.16713},
  year={2025}
}

@article{zhi2024closed,
  title={Closed-loop open-vocabulary mobile manipulation with gpt-4v},
  author={Zhi, Peiyuan and Zhang, Zhiyuan and Zhao, Yu and Han, Muzhi and Zhang, Zeyu and Li, Zhitian and Jiao, Ziyuan and Jia, Baoxiong and Huang, Siyuan},
  journal={arXiv preprint arXiv:2404.10220},
  year={2024}
}

@inproceedings{raman2024cape,
  title={Cape: Corrective actions from precondition errors using large language models},
  author={Raman, Shreyas Sundara and Cohen, Vanya and Idrees, Ifrah and Rosen, Eric and Mooney, Raymond and Tellex, Stefanie and Paulius, David},
  booktitle={2024 IEEE International Conference on Robotics and Automation (ICRA)},
  pages={14070--14077},
  year={2024},
  organization={IEEE}
}

@article{styrud2024automatic,
  title={Automatic behavior tree expansion with llms for robotic manipulation},
  author={Styrud, Jonathan and Iovino, Matteo and Norrl{\"o}f, Mikael and Bj{\"o}rkman, M{\aa}rten and Smith, Christian},
  journal={arXiv preprint arXiv:2409.13356},
  year={2024}
}

@article{quoc2025vision,
  title={A Vision-Language Foundation Model for Leaf Disease Identification},
  author={Quoc, Khang Nguyen and Thu, Lan Le Thi and Quach, Luyl-Da},
  journal={arXiv preprint arXiv:2505.07019},
  year={2025}
}

@article{nawaz2024agriclip,
  title={AgriCLIP: Adapting CLIP for agriculture and livestock via domain-specialized cross-model alignment},
  author={Nawaz, Umair and Awais, Muhammad and Gani, Hanan and Naseer, Muzammal and Khan, Fahad and Khan, Salman and Anwer, Rao Muhammad},
  journal={arXiv preprint arXiv:2410.01407},
  year={2024}
}

@article{shinoda2025agrobench,
  title={AgroBench: Vision-Language Model Benchmark in Agriculture},
  author={Shinoda, Risa and Inoue, Nakamasa and Kataoka, Hirokatsu and Onishi, Masaki and Ushiku, Yoshitaka},
  journal={arXiv preprint arXiv:2507.20519},
  year={2025}
}

@article{zhang2024visual,
  title={Visual large language model for wheat disease diagnosis in the wild},
  author={Zhang, Kunpeng and Ma, Li and Cui, Beibei and Li, Xin and Zhang, Boqiang and Xie, Na},
  journal={Computers and Electronics in Agriculture},
  volume={227},
  pages={109587},
  year={2024},
  publisher={Elsevier}
}

@inproceedings{awais2025agrogpt,
  title={Agrogpt: Efficient agricultural vision-language model with expert tuning},
  author={Awais, Muhammad and Alharthi, Ali Husain Salem Abdulla and Kumar, Amandeep and Cholakkal, Hisham and Anwer, Rao Muhammad},
  booktitle={2025 IEEE/CVF Winter Conference on Applications of Computer Vision (WACV)},
  pages={5687--5696},
  year={2025},
  organization={IEEE}
}

@article{zhou2024few,
  title={Few-shot image classification of crop diseases based on vision--language models},
  author={Zhou, Yueyue and Yan, Hongping and Ding, Kun and Cai, Tingting and Zhang, Yan},
  journal={Sensors},
  volume={24},
  number={18},
  pages={6109},
  year={2024},
  publisher={MDPI}
}

@article{yu2024framework,
  title={A Framework for Agricultural Intelligent Analysis Based on a Visual Language Large Model},
  author={Yu, Piaofang and Lin, Bo},
  journal={Applied Sciences},
  volume={14},
  number={18},
  pages={8350},
  year={2024},
  publisher={MDPI}
}

@inproceedings{goetting2024end,
  title={End-to-end navigation with vlms: Transforming spatial reasoning into question-answering},
  author={Goetting, Dylan and Singh, Himanshu Gaurav and Loquercio, Antonio},
  booktitle={Workshop on Language and Robot Learning: Language as an Interface},
  year={2024}
}

@article{batra2020objectnav,
  title={Objectnav revisited: On evaluation of embodied agents navigating to objects},
  author={Batra, Dhruv and Gokaslan, Aaron and Kembhavi, Aniruddha and Maksymets, Oleksandr and Mottaghi, Roozbeh and Savva, Manolis and Toshev, Alexander and Wijmans, Erik},
  journal={arXiv preprint arXiv:2006.13171},
  year={2020}
}

@article{kaiser2012high,
  title={High tunnel tomatoes},
  author={Kaiser, Cheryl and Ernst, Matt},
  journal={University of Kentucky College of Agriculture, Food and Environment Cooperative Extension Service, Lexington},
  year={2012}
}

@article{waterer2003yields,
  title={Yields and economics of high tunnels for production of warm-season vegetable crops},
  author={Waterer, Doug},
  journal={HortTechnology},
  volume={13},
  number={2},
  pages={339--343},
  year={2003},
  publisher={American Society for Horticultural Science}
}

@article{groot2024overconfidence,
  title={Overconfidence is key: Verbalized uncertainty evaluation in large language and vision-language models},
  author={Groot, Tobias and Valdenegro-Toro, Matias},
  journal={arXiv preprint arXiv:2405.02917},
  year={2024}
}

@inproceedings{zhou2022detecting,
  title={Detecting twenty-thousand classes using image-level supervision},
  author={Zhou, Xingyi and Girdhar, Rohit and Joulin, Armand and Kr{\"a}henb{\"u}hl, Philipp and Misra, Ishan},
  booktitle={European conference on computer vision},
  pages={350--368},
  year={2022},
  organization={Springer}
}

@inproceedings{radford2021learning,
  title={Learning transferable visual models from natural language supervision},
  author={Radford, Alec and Kim, Jong Wook and Hallacy, Chris and Ramesh, Aditya and Goh, Gabriel and Agarwal, Sandhini and Sastry, Girish and Askell, Amanda and Mishkin, Pamela and Clark, Jack and others},
  booktitle={International conference on machine learning},
  pages={8748--8763},
  year={2021},
  organization={PmLR}
}

@misc{ros-move_base, author = {Hershberger, E. and Madrigal, E. and Gerkey, B.}, title = {move\_base}, year = {2020}, publisher = {GitHub}, journal = {GitHub repository}, howpublished = {\url{https://wiki.ros.org/move_base}}, note = {Accessed: 2026-03-17} }

@article{coleman2014reducing, title={Reducing the Barrier to Entry of Complex Robotic Software: a MoveIt! Case Study}, author={Coleman, David and {\c{S}}ucan, Ioan A and Chitta, Sachin and Correll, Nikolaus}, journal={Journal of Software Engineering for Robotics}, volume={5}, number={1}, pages={3--16}, year={2014}, doi={10.6092/JOSER_2014_05_01_p3} }

@article{brohan2022rt,
  title={Rt-1: Robotics transformer for real-world control at scale},
  author={Brohan, Anthony and Brown, Noah and Carbajal, Justice and Chebotar, Yevgen and Dabis, Joseph and Finn, Chelsea and Gopalakrishnan, Keerthana and Hausman, Karol and Herzog, Alex and Hsu, Jasmine and others},
  journal={arXiv preprint arXiv:2212.06817},
  year={2022}
}

@inproceedings{zitkovich2023rt,
  title={Rt-2: Vision-language-action models transfer web knowledge to robotic control},
  author={Zitkovich, Brianna and Yu, Tianhe and Xu, Sichun and Xu, Peng and Xiao, Ted and Xia, Fei and Wu, Jialin and Wohlhart, Paul and Welker, Stefan and Wahid, Ayzaan and others},
  booktitle={Conference on Robot Learning},
  pages={2165--2183},
  year={2023},
  organization={PMLR}
}

@article{kim2024openvla,
  title={Openvla: An open-source vision-language-action model},
  author={Kim, Moo Jin and Pertsch, Karl and Karamcheti, Siddharth and Xiao, Ted and Balakrishna, Ashwin and Nair, Suraj and Rafailov, Rafael and Foster, Ethan and Lam, Grace and Sanketi, Pannag and others},
  journal={arXiv preprint arXiv:2406.09246},
  year={2024}
}

@article{black2024pi_0,
  title={A Vision-Language-Action Flow Model for General Robot Control},
  author={Black, Kevin and Brown, Noah and Driess, Danny and Esmail, Adnan and Equi, Michael and Finn, Chelsea and Fusai, Niccolo and Groom, Lachy and Hausman, Karol and Ichter, Brian and others},
  journal={arXiv preprint arXiv:2410.24164},
  year={2024}
}

@inproceedings{weerakoon2025behav,
  title={Behav: Behavioral rule guided autonomy using vlms for robot navigation in outdoor scenes},
  author={Weerakoon, Kasun and Elnoor, Mohamed and Seneviratne, Gershom and Rajagopal, Vignesh and Arul, Senthil Hariharan and Liang, Jing and Jaffar, Mohamed Khalid M and Manocha, Dinesh},
  booktitle={2025 IEEE International Conference on Robotics and Automation (ICRA)},
  pages={7044--7051},
  year={2025},
  organization={IEEE}
}

@article{gummadi2025zest,
  title={ZeST: an LLM-based Zero-Shot Traversability Navigation for Unknown Environments},
  author={Gummadi, Shreya and Gasparino, Mateus V and Capezzuto, Gianluca and Becker, Marcelo and Chowdhary, Girish},
  journal={arXiv preprint arXiv:2508.19131},
  year={2025}
}

@inproceedings{sathyamoorthy2024convoi,
  title={Convoi: Context-aware navigation using vision language models in outdoor and indoor environments},
  author={Sathyamoorthy, Adarsh Jagan and Weerakoon, Kasun and Elnoor, Mohamed and Zore, Anuj and Ichter, Brian and Xia, Fei and Tan, Jie and Yu, Wenhao and Manocha, Dinesh},
  booktitle={2024 IEEE/RSJ International Conference on Intelligent Robots and Systems (IROS)},
  pages={13837--13844},
  year={2024},
  organization={IEEE}
}

@inproceedings{elnoor2025vlm,
  title={VLM-GroNav: Robot Navigation Using Physically Grounded Vision-Language Models in Outdoor Environments},
  author={Elnoor, Mohamed and Weerakoon, Kasun and Seneviratne, Gershom and Xian, Ruiqi and Guan, Tianrui and Jaffar, Mohamed Khalid M and Rajagopal, Vignesh and Manocha, Dinesh},
  booktitle={2025 IEEE International Conference on Robotics and Automation (ICRA)},
  pages={2391--2398},
  year={2025},
  organization={IEEE}
}

@article{song2024vlm,
  title={Vlm-social-nav: Socially aware robot navigation through scoring using vision-language models},
  author={Song, Daeun and Liang, Jing and Payandeh, Amirreza and Raj, Amir Hossain and Xiao, Xuesu and Manocha, Dinesh},
  journal={IEEE Robotics and Automation Letters},
  volume={10},
  number={1},
  pages={508--515},
  year={2024},
  publisher={IEEE}
}

@inproceedings{cheng2024navila,
title = {NaVILA: Legged Robot Vision-Language-Action Model for Navigation},
    author = {Cheng, An-Chieh and Ji, Yandong and Yang, Zhaojing and Zou, Xueyan and Kautz, Jan and Biyik, Erdem and Yin,
    Hongxu and Liu, Sifei and Wang, Xiaolong},
    booktitle = {RSS},
    year = {2025},
}

@article{chiang2024mobility,
  title={Mobility vla: Multimodal instruction navigation with long-context vlms and topological graphs},
  author={Chiang, Hao-Tien Lewis and Xu, Zhuo and Fu, Zipeng and Jacob, Mithun George and Zhang, Tingnan and Lee, Tsang-Wei Edward and Yu, Wenhao and Schenck, Connor and Rendleman, David and Shah, Dhruv and others},
  journal={arXiv preprint arXiv:2407.07775},
  year={2024}
}

@article{zhao2025agrivln,
  title={AgriVLN: Vision-and-Language Navigation for Agricultural Robots},
  author={Zhao, Xiaobei and Lyu, Xingqi and Li, Xiang},
  journal={arXiv preprint arXiv:2508.07406},
  year={2025}
}
\end{document}